\documentclass[journal,compsoc]{IEEEtran}

\usepackage{mathbbol}
\usepackage{graphicx}
\usepackage{hyperref}
\usepackage{textcomp}
\usepackage[]{algorithm}
\usepackage[]{algpseudocode}
\usepackage[caption=false]{subfig}
\usepackage{amsmath}
\usepackage{diagbox}
\usepackage{graphicx}
\usepackage[utf8]{inputenc}
\usepackage{bookmark}
\usepackage{float}
\usepackage[capitalize]{cleveref}
\usepackage{epstopdf}
\usepackage{enumitem}
\usepackage{caption}    
\usepackage{enumitem}

\usepackage[backend=biber,
date=year, 
style=ieee]{biblatex}

\usepackage{color, colortbl}
\newcommand{\arxiv}[1]{#1} 

\begin{document}

	\title{Learning to Evolve}

	\author{

		Jan Schuchardt, Vladimir Golkov, Daniel Cremers
		\thanks{Authors are with the Department of Informatics, Technical University of Munich, Germany.
			e-mail: \url{jan.schuchardt@tum.de}, \url{vladimir.golkov@tum.de}, \url{cremers@tum.de}}
	}

	\markboth{{Schuchardt \MakeLowercase{\textit{et al.}}: Learning to Evolve}}
	{{Schuchardt \MakeLowercase{\textit{et al.}}: Learning to Evolve}}

	\maketitle

	\begin{abstract}
		Evolution and learning are two of the fundamental mechanisms by which life adapts in order to survive and to transcend limitations. These biological phenomena inspired successful computational methods such as evolutionary algorithms and deep learning. Evolution relies on random mutations and on random genetic recombination. Here we show that \emph{learning to evolve}, i.e.~learning to mutate and recombine better than at random, improves the result of evolution in terms of fitness increase per generation and even in terms of attainable fitness. We use deep reinforcement learning to learn to dynamically adjust the strategy of evolutionary algorithms to varying circumstances. Our methods outperform classical evolutionary algorithms on combinatorial and continuous optimization problems.
	\end{abstract}

	\IEEEpeerreviewmaketitle
	
	\setcounter{tocdepth}{4}\tableofcontents

	\section{Introduction}

	\IEEEPARstart{M}{ost} problems in engineering and the natural sciences can be formulated as optimization problems. Evolutionary computation is inspired by the powerful mechanisms of natural evolution. While other methods might get easily stuck when optimizing rugged objective functions, evolutionary algorithms can escape local optima, explore the solution space through random mutation and combine favorable features of different solutions through crossover, all while being simple to implement and parallelize.
	
	Consequently, evolutionary algorithms have been applied to a range of engineering problems, from designing steel-beams~\autocite{Kameshki2001} and antennas for space-missions~\autocite{Lohna}, to more large-scale problems, like the design of wind parks~\autocite{Mosetti1994}, water supply networks, or smart energy grids. 
	
	Evolution and learning are two optimization frameworks that in living systems work at different scales with different advantages. Appropriate combinations of the two provide complementarity, and are a crucial part of the success of living systems. Here we propose new combinations of these optimization principles.
	
	We propose using deep reinforcement learning to dynamically control the parameters of evolutionary algorithms. The goal is finding better solutions to hard optimization problems and facilitating the application of evolutionary algorithms.
	
	Our \emph{deep learning for evolutionary algorithms} is not to be confused with \emph{evolutionary algorithms for deep learning}, such as neuroevolution~\autocite{Stanley2019} or population-based training~\autocite{alphastarblog}.
	
	This section provides a brief explanation of the used terms and definitions. \cref{related-work} provides an overview of previous work dedicated to enhancing evolutionary algorithms through reinforcement learning. \cref{methods} explains how we aim to do away with the shortcomings of previous work, as well as the experimental setup used to provide initial evidence of the feasibility of our approach. \cref{results} is dedicated to the results of our experiments and their discussion. \cref{conclusions} provides high-level conclusions to our experimental results.
	
	\subsection{Evolutionary Computation}\label{evolutionary-computation}
	Evolutionary computation is an umbrella term for optimization methods inspired by Darwinian evolutionary theory. In natural evolution, individuals strive for survival and reproduction in a competitive environment. Those with more favorable traits, acquired through inheritance or mutation, have a higher chance of succeeding.
	
	\begin{figure*}[]
		\centering
		\includegraphics[width=0.95\linewidth]{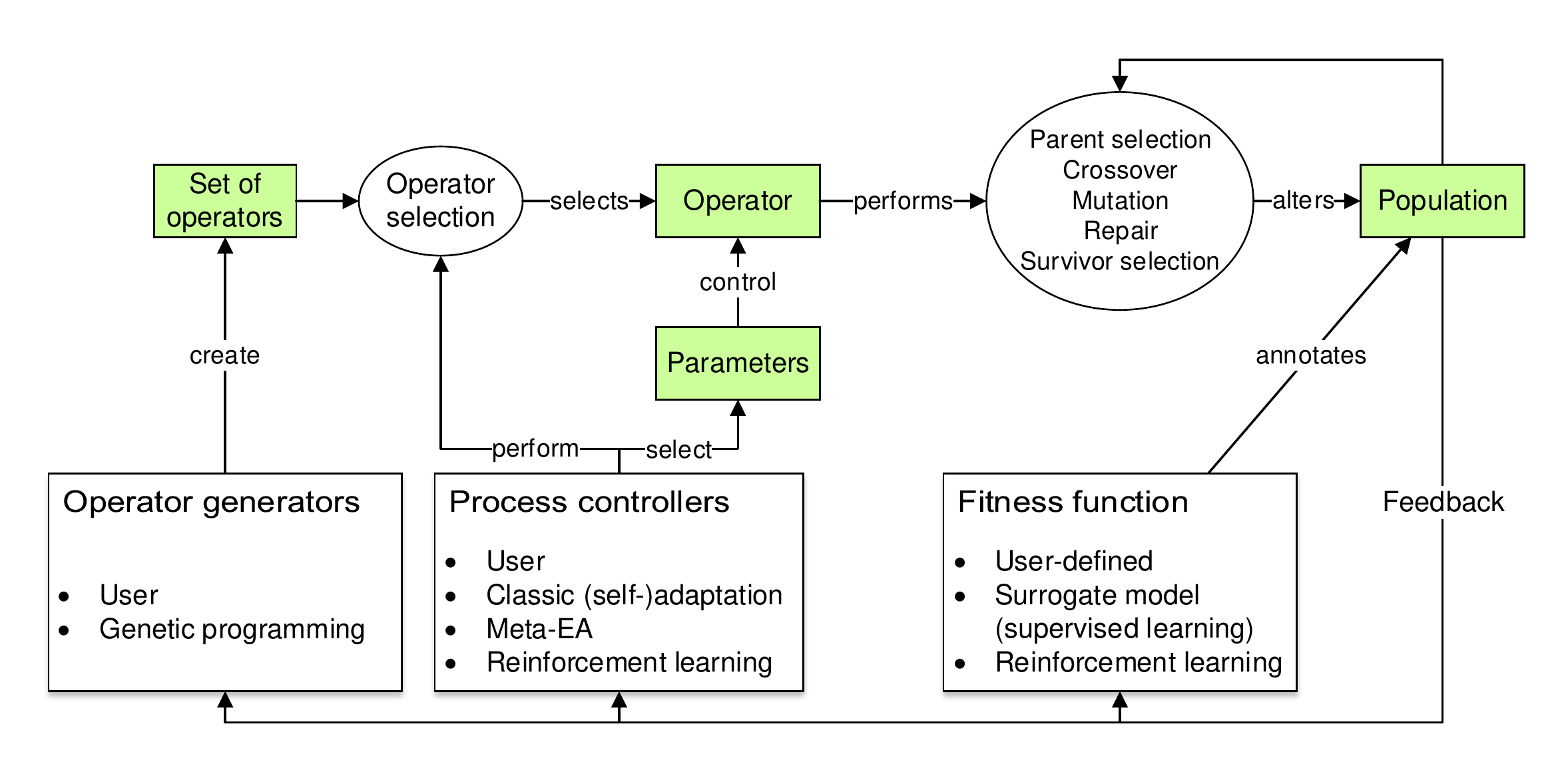}
		\caption{Data flow in evolutionary algorithms.
		Operator generation, process control and fitness estimation can either be entirely predefined by the user, or follow some algorithmic approach and receive feedback from the population. We propose using deep reinforcement learning instead of classical process controllers or fitness estimators.}
		\label{ea-pipeline}
	\end{figure*}
	
	Evolutionary algorithms are specific realizations of the concept of evolutionary computation.
	Evolutionary algorithms solve computational problems by managing a set (\emph{population}) of \emph{individuals}. Each individual encodes a candidate solution to the computational problem in its \emph{genome}.
	To explore the solution space, \emph{offspring} is generated from the \emph{parent population} through \emph{recombination operators} that combine properties of the parents. Additionally, \emph{mutation operators} are applied to introduce random variations with the goal of enhancing exploration and preventing premature convergence. A new population is created by selecting a set of individuals from the parent population and from the offspring.
	This process of \emph{recombination}, \emph{mutation} and \emph{survivor selection} comprises one \emph{generation} and is repeated multiple times throughout a run of the algorithm.
	To guide the process towards better solutions, evolutionary pressure is applied through a \emph{fitness function}. Fitter individuals are given a higher chance of reproducing, surviving, or both.	
	Due to its well parallelizable design and their suitability for solving high-dimensional problems with a complex fitness landscape, evolutionary computation is a valuable tool in engineering applications or other domains where classical optimization methods fail or no efficient exact solver is available.
	\cref{ea-pipeline} shows the data flow of evolutionary algorithms, and the role we propose therein for reinforcement learning.
	
	\subsection{Adaptation in Evolutionary Computation}

	A key problem in the application of evolutionary algorithms is selecting evolution parameters. Even simple implementations have a considerable number of parameters.
	
	The choice of parameter values has a considerable impact on the performance of an evolutionary algorithm for different problems and even different problem instances. Furthermore, utilizing fixed parameters over the course of all generations can be sub-optimal, as different stages of the search process might have different requirements.
	
	To account for this, it is desirable for evolutionary algorithms to be adaptive. In this context, \emph{adaptation} refers to dynamic control of evolution parameters (not to be confused with the biological term \emph{adaptation}).

	\label{levels-of-adaptation}
	The following taxonomy, taken from Ref.~\autocite{Hinterdinga}, describes the different levels on which adaptation can be used in evolutionary algorithms.
	\begin{enumerate}
		\itemsep0.5em
		\item \emph{Environment-Level Adaptation} changes the way in which individuals are evaluated by the environment, for example by altering the fitness function.
		\item \emph{Population-Level Adaptation} modifies parameters that affect the entirety or some subset of the population, for example by changing the population size. 
		\item \emph{Individual-Level Adaptation} makes parameter choices for specific individuals in the population, for example by increasing the mutation probability of individuals with a low fitness value.
		\item \emph{Component-Level Adaptation} changes parameters that are specific to a certain part of an individual's genome, for example by managing per-gene mutation probabilites.
	\end{enumerate}
	
	In \cref{subsection-adaptation-methods}, we propose adaptation methods for each of these levels of adaptation.
	
	\subsection{Deep Reinforcement Learning}
	Artificial neural networks are a popular machine learning technique and connectionist model, inspired by neurobiology.
	
	A single neuron is parameterized by weights that model how its inputs relate to its output. By combining multiple neural layers that perform (nonlinear) data transformations, highly complicated functions can be approximated. In order to model a mapping with desired properties, the network weights are modified to minimize a loss function. This minimization is usually implemented using some form of gradient descent.
	
	Artificial neural networks have seen a surge in popularity in the past few years and have been successfully applied in a variety of fields, like computer vision, natural language processing, biology, medicine, finance, marketing and others.

	Reinforcement learning is an area of artificial intelligence concerned with training a learning agent by providing rewards for the actions it takes in different states of its environment. The ultimate goal is for the agent to follow a policy that maximizes these rewards.
	
	The key limiting factor in the application of older reinforcement learning methods is the complexity of representing policies for large state and action spaces. Deep reinforcement learning is the idea of using artificial neural networks as function approximators that replace tabular or other representations used in classic reinforcement learning algorithms. One additional benefit of deep reinforcement learning is that it made the use of reinforcement learning for continuous control feasible (see, for example~\autocite{Lillicrap2015}).

	\section{Related Work}\label{related-work}

	To our knowledge, there has been no previous work on the application of deep reinforcement learning to evolutionary computation.
	However, there have been several publications on using classic reinforcement learning techniques for adaptation in evolutionary algorithms.
	
	Most previous work has been concerned with population-level adaptation. In 2002, Müller et al.
	enhanced a $(1+1)$ evolution strategy (i.e.~an evolutionary algorithm for continuous optimization with population size 1) by controlling the step-size (i.e.~standard deviation)  through reinforcement learning~\autocite{Mullera}. Later work~\autocite{Eibena,Karafotias2014} extended the use of reinforcement learning to the simultaneous control of multiple numerical evoluton parameters. Aside from this, reinforcement learning has also been used to dynamically select from a set of available evolutionary operators \autocite{Pettinger2,Buzdalova2014}. Techniques for multi-armed bandits (i.e.~reinforcement learning with a single state) have also been utilized for this purpose, both in single-objective~\autocite{DaCosta2008} and multi-objective optimization~\autocite{Li2014}.
	
	Reinforcement learning has also been successfully applied to environment-level adaptation. In their 2011 paper~\autocite{Afanasyeva2011}, Afanasyeva and Buzdalov used reinforcement learning to select from a set of handcrafted auxiliary fitness functions that can be added to the main objective function, in order to reshape the fitness landscape. This approach was later expanded on to deal with non-stationary problems, in which the objective function changes over time~\autocite{Petrova2014}.
	
	As individual- and component-level adaptation requires larger action spaces, with which classic reinforcement learning algorithms struggle, there has been little research into learning such strategies through reinforcement learning. The only paper we were able to find used reinforcement learning to control two numerical evolution parameters per individual in the local search strategy of a memetic algorithm~\autocite{Bhowmik2012} (i.e.~a combination of an evolutionary algorithm with a local search strategy). To our knowledge, there has been no previous work on reinforcement-learning based component-level adaptation.
	
	A limiting factor in all of these approaches is that they employ reinforcement learning methods like Q-Learning~\autocite{Watkins1992} that represent the policy of the learning agent in a discretized fashion. Practical application of these older reinforcement learning methods is limited to learning low-dimensional mappings with a small number of state-action pairs.
	
	Consequently, only a small, coarsely discretized subset of the potentially useful information about the optimization problem and the state of the evolutionary algorithms is used. Likewise, the action space is discretized coarsely, even though many of the controlled parameters are continuous in nature. There has been an attempt to address the problem of action space discretization by dynamically adapting the discretization bins~\autocite{Rost2016}, but this does resolve the problem that the underlying reinforcement learning algorithm is ill-suited to continuous control.
	
	It should also be noted that all aforementioned work is concerned with learning  on the fly (i.e.~during the execution of an evolutionary algorithm) for a specific problem instance, whereas our method is designed to learn over the course of multiple runs of the evolutionary algorithm, as explained in the next section.

	\section{Methods}\label{methods}
	To do away with the limitations of older approaches (see the end of \cref{related-work}), we propose using deep reinforcement learning to learn adaptation strategies for evolutionary algorithms.
	
	The novelties of our approach include:
	\begin{itemize}
		\item Learning adaptation strategies for an entire problem class, instead of optimizing for a specific problem instance
		\item Using more information about problem instances and the state of the evolutionary algorithm
		\item Utilizing modern deep reinforcement learning techniques in order to

		operate in large, continuous state and action spaces.
		
		This allows us to:
		\begin{itemize}
			\item Learn complex adaptation strategies
			\item Entirely replace hand-crafted components of an evolutionary algorithm (e.g.~parent selection) with learned strategies
		\end{itemize}
	\end{itemize}
	
	While many other use cases are possible (see~\cref{conclusions}), we limit ourselves to learning adaptation that generalizes to previously unseen problem instances, using only a limited number of instances of the same problem class for training, and always running the evolutionary algorithms for a fixed number of generations. Within these constraints, we consider two distinct use cases:
	
	\begin{enumerate}
		\item The time/resources for training are large. In this case, the user can account for possible instabilities of the training process by selecting the best out of multiple trained agents.
		\item The time/resources for training are limited, only allowing for the training of one or very few agents. In this case, it is important that the average performance of trained agents is high and the variance in performance among them is low, so that the user is likely to arrive at a good solution within the limitations of this use case.
	\end{enumerate}

	The rest of the Methods section is structured as follows. We first explain our used reinforcement learning approach (\cref{reinforcement-learning-methods}) and reward function (\cref{reward-calculation}). We then define three benchmark problem sets (\cref{benchmark-problems}) and basic evolutionary algorithms that can be used to optimize these problems (\cref{baseline-algorithms}). Next, we propose different trainable adaptation methods to enhance these evolutionary algorithms (\cref{subsection-adaptation-methods}) and specify the neural network architecture used for performing the underlying calculations (\cref{network-architecture}). Finally, we define the experimental setup and performance metrics used for evaluating the different proposed adaptation methods (\cref{experimental-setup}).

	We will release the code at \url{https://github.com/jan-schuchardt/learning-to-evolve}
	
	\subsection{Choice of Reinforcement Learning Algorithm}\label{reinforcement-learning-methods}
	To allow us to perform both discrete and continuous actions -- depending on the application -- we propose the use of so-called stochastic policy gradient methods, which take actions  by sampling from a probability distribution, parameterized by a neural network. Most state-of-the-art deep reinforcement learning algorithms fall into this category.

	In this section, we first provide a more formal definition of reinforcement learning (\cref{reinforcment-learning}), before explaining the specifics of our used reinforcement learning algorithm (\cref{ppo}).
	
	\subsubsection{Basics of Reinforcement Learning}\label{reinforcment-learning}
	Reinforcement learning is a field of study concerned with training intelligent agents through rewards or penalties, based on actions taken in an environment.
	
	Reinforcement learning problems are typically specified as a Markov decision process, defined by:
	\begin{itemize}
		\item a set $S$ of states,
		\item an initial state $s_0$ or a probability distribution $p(S_{0})$ over a set $S_0$ of initial states,
		\item a set $A$ of actions,
		\item the transition function $\mathcal{P}_a(s^{\prime} {\mid} s)$, which describes the probability of reaching state $s^{\prime}$ from state $s$ by taking action $a$,
		\item the reward function $\mathcal{R}_a(s^{\prime}, s)$ which assigns scalar rewards (larger is better) to a state transition,
		\item the Markov property $P(s_{t+1} {\mid} s_t, s_{t-1}, \dots, s_0) = P(s_{t+1} {\mid} s_t)$, meaning that the state transitions at time $t$ are independent of the prior sequence of states.
	\end{itemize}
	
	The goal of reinforcement learning is to learn a policy $\mathcal{\pi}:S, A \to [0, 1]$ that describes a probability distribution over actions, given a state. The learning has to be achieved solely based on the observed rewards and state transitions, without prior knowledge of the environment. The policy should maximize the expected value of some reward-based return function $R$. A typical choice is the discounted sum of accumulated rewards:
	\begin{displaymath}\label{discounted-sum-of-rewards}
	R= \sum_{t=0}^{\infty}{\gamma^{t} r_{t}},
	\end{displaymath}
	with $r$ being the sequence of received rewards, and $\gamma \in [0, 1)$ being a discount factor that ensures convergence of the series. A smaller $\gamma$ means that short-term rewards are favored over long-term rewards.
	
	\subsubsection{Proximal Policy Optimization}\label{ppo}
	Proximal policy optimization~\autocite{Schulman2017} is a stochastic policy gradient method that aims to enhance training stability by using trust-region optimization of the policy.
	
	In stochastic policy gradient methods, the gradient of expected future rewards with respect to the parameters $\theta$ of a stochastic policy $ \pi_{\theta}$ is used for learning. Each state is mapped to a probability distribution over actions. An action is selected by sampling from this probability distribution.

	Proximal policy optimization uses the following clipped loss function for training:
	\begin{equation}
	\mathcal{L}_\mathrm{clip} = - \mathbb{E}[\min(r_t(\theta)\hat{A_t}, \mathrm{clip}(r_t(\theta), [ 1 - \varepsilon, 1 + \varepsilon ])\hat{A_t} ) ],
	\end{equation}
	\begin{equation}
	\textnormal{with } r_t(\theta) = \frac{\pi_{\theta}( a_t {\mid} s_t)} { \pi_{\theta_{\mathrm{old}}} (a_t {\mid} s_t)},
	\end{equation}
	where $\mathbb{E}$ is the average over a set of training samples,  $\pi_{\theta}(a_t {\mid} s_t)$ is the probability of performing action $a_t$ in state $s_t$ under the probability distribution described by the policy $\pi_{\theta}$,  $\theta_{\mathrm{old}}$ are the parameters of the policy during collection of the training samples and $\theta$ are the parameters of the policy as it undergoes optimization. The advantage estimator $\hat{A_t}$  describes how much higher than expected the reward for following action $a_t$ at time $t$ was. The clipping function  $\mathrm{clip}(r_t(\theta), [1 - \varepsilon, 1 + \varepsilon])$ maps $r_t(\theta)$ to the interval $[1 - \varepsilon, 1 + \varepsilon]$, i.e.~$\mathrm{clip}(r_t(\theta), [1 - \varepsilon, 1 + \varepsilon]) := \min\{\max\{r_t(\theta), 1 - \varepsilon\}, 1 + \varepsilon\}$. The benefit of this formulation is that it does not encourage increasing the probability of an advantageous $a_t$ ($\hat{A_t} > 0)$ or decreasing the probability of a worse-than-expected $a_t$ ($\hat{A_t} < 0$) by more than $\varepsilon$, thus stabilizing the learning process and allowing for training samples to be reused without perturbing the policy, which increases sample efficiency.

	\begin{algorithm}[htb]
		\begin{algorithmic}
			\For{iteration=1,2, \dots ,\#iterations}
			\For{problem\_instance=1,2, \dots , $K$}
			\For{actor=1,2, \dots ,$N$}
			\State{Run policy $\pi_{\theta_\mathrm{old}}$ for $T$ timesteps}
			\State{Calculate $\hat{A_1}, \hat{A_2} \dots , \hat{A_T}$}
			\State{Store training samples}
			\EndFor
			\EndFor
			\For{epoch=1,2, \dots ,\#epochs}
			\State{Optimize clipped loss on samples w.r.t.\ $\theta$, using minibatch size $M \leq KNT$}
			\EndFor
			\State{$\pi_{\theta_{old}} \gets \pi_{\theta}$}
			\State{Discard training samples}
			\EndFor
		\end{algorithmic}
		\caption{Proximal policy optimization for multiple problem instances}
		\label{ppo-algorithm}
	\end{algorithm}
	
	\cref{ppo-algorithm} specifies how we use proximal policy optimization to optimize a policy for multiple problem instances. In each training iteration, the evolutionary algorithm is applied to each problem instances for a fixed number of times, in order to gather samples for subsequent training.
	
	\paragraph{Advantage Estimation}\label{generalized-advantage-estimation}

	We use generalized advantage estimation (see~\autocite{Schulman2015a}) for calculating the advantage estimate $\hat{A}$ while ensuring a good trade-off between variance and bias.
	
	Assuming an estimator $\hat{V}(s_t)$ (\emph{value function}) of the discounted future rewards $V(s_t) = \sum_{i=0}^{\infty} \gamma^i r_{t+i}$, the advantage for a \emph{trajectory} (i.e.~sequence of state transitions, actions and rewards) of length $T$ is calculated as an exponential moving average over temporal differences:
	
	\begin{equation}
	\hat{A_t} =  \sum_{i=0}^{T-t+1}(\gamma \lambda)^i \delta_{t+i},
	\end{equation}
	\begin{equation}
	\textnormal{with } \delta_t = \gamma \hat{V}(s_{t+1}) + r_t - \hat{V}_t(s_t), 
	\end{equation}
		where the parameter $\lambda \in [0, 1]$ controls the trade-off between variance and bias of the advantage estimate.
		With higher $\lambda$, the sequence of rewards is given a higher weight, thus reducing the bias caused by the estimate $\hat{V}(s_{t})$.
		However, the variance of the estimate increases with $T$, due to the randomness of the underlying Markov decision process.

	\paragraph{Time-Awareness}\label{time-awareness}

	In our approach, the evolutionary algorithm is run for a limited number of generations. Consequently, learned adaptation methods should try to maximize the fitness within this given time frame.
	
	For simplicity, we treat the entire run of the evolutionary algorithm as a single episode of length $T$.
	To account for the time-limited nature of the environment, we make the following adjustments, based on \autocite{Pardo2017}:
	\begin{enumerate}
		\item \label{limited-time-evolutionary} We enforce $\hat{V}(s_T) = 0$ in the generalized advantage estimation, as no further rewards can be gathered after the episode has ended.
		\item \label{time-encoding} We add a relative encoding of the remaining number of generations, $(T - t) / T$, to the state. This way, the policy can account for optimization problems in which the potential for gathering rewards might be considerably higher at the earlier state and adapt its behavior accordingly. By using a relative encoding, we scale $T-t$ into the $[0, 1]$ range, with the goal of better generalization when applying the evolutionary algorithm under varying $T$.
	\end{enumerate}

	\paragraph{Actor-Critic Framework}\label{actor-critic}

	There are a variety of ways to calculate the value approximator $\hat{V}(s_t)$, one of which is approximating $V$ through a neural network with parameters $\theta_c$. We optimize  $\hat{V}_{\theta_c}$ by minimizing
	\begin{equation}
	\mathcal{L}_V = \mathbb{E}\left[ \left\lVert \hat{V}_{\theta_c}(s_t) - \sum_{i=0}^{T-t} \gamma^i r_{t+i} \right\rVert^2 \right].
	\end{equation}

	This neural network is typically referred to as a \emph{critic}, rating the actions taken by the \emph{actor} $\pi_{\theta}$. Since both operate in the same environment, it is common practice to merge them into one network and only keep two separate output layers, so that they can operate on shared lower-level features.
	In this case, the loss for the entire network is $\mathcal{L}_\mathrm{clip} + \alpha_v \mathcal{L}_V$, with the hyper-parameter $\alpha_v$ controlling the ratio between the actor and critic losses.

	\paragraph{Entropy-Based Exploration}\label{entropy-exploration}
	In order to learn a good policy and avoid bad local optima, it is vital to explore a variety of actions and states. To this end, the negative information-theoretical entropy $S[\pi_{\theta}]$ of $\pi_{\theta}$ can be added to the loss function~\autocite{Mnih2016}. By minimizing this term (maximizing the entropy), actions are taken with less certainty, thus discouraging premature convergence to local optima. This leads to the complete loss function
	$
	\mathcal{L} = \mathcal{L}_{\mathrm{clip}} + \alpha_v \mathcal{L}_V + \alpha_e S[\pi_{\theta}],
	$
	with the exploration-controlling coefficient $\alpha_e$.

	\subsubsection{Reward Calculation}
	\label{reward-calculation}

	The goal of a reinforcement learning algorithm is to maximize rewards over the course of multiple state transitions (see \cref{reinforcment-learning}), 
	while the goal of an evolutionary algorithm is to find a solution of maximum fitness over the course of multiple generations (see \cref{evolutionary-computation}). To unify these two goals, we associate the state $s_t$ with the population of the evolutionary algorithm in generation $t$.
	
	Let $f_\mathrm{max}(s_t)$ be a function that returns the fitness value of the fittest individual in the population associated with $s_t$.
	On a generation-to-generation level, the goal of finding a solution of maximum fitness then translates to maximizing the ratio $f_\mathrm{max}(s_{t+1})/f_\mathrm{max}(s_t)$.

	Multiple smaller improvements should have the same effect as one large improvement that leads to the same final solution. We therefore define the reward function as
	\begin{equation}
	\mathcal{R}_a(s_{t}, s_{t+1}) = \alpha_r \log_{10} \frac{f_\mathrm{max}(s_{t+1})}{f_\mathrm{max}(s_t)},
	\end{equation}
	assuming positive fitness functions. We use the coefficient $\alpha_r$ to scale rewards approximately into the range $[-1, 1]$. The logarithm is taken, so that the sum of rewards over one run of the evolutionary algorithm equals the logarithm of the ratio between the initial and terminal fitness, $\alpha_r \log_{10} \frac{f_\mathrm{max}(s_{T})}{f_\mathrm{max}(s_0)}$. 
	
	\subsection{Benchmark Problems}\label{benchmark-problems}
	
	Three different problem classes are used to investigate the usefulness of different reinforcement learning adaptation mechanisms. We use the 0-1 knapsack problem and the traveling salesman problem as examples for combinatorial optimization, and a set of two-dimensional objective functions as examples for continuous optimization.
	As explained in the beginning of \cref{methods}, we only use a limited number of problem instances for training.
	
	This section gives a brief explanation of the different optimization problems and how we define their respective fitness functions.
	
	\subsubsection{0-1 Knapsack Problem}\label{0-1-knapsack}
	An instance of the 0-1 Knapsack Problem is defined by a weight limit $w_{\mathrm{max}}$ and a set $I$ of $n$ items: $I = {\{(w_i, v_i) \mid w, v \in \mathbb{R}, i \in [n]\}}$, with weights $w$ and values $v$.
	The optimization objective is
	\begin{equation}
	\mathop{\textrm{max}}_{S \subseteq I} \sum_{ (w, v) \in S}{v}  \quad \textrm{subject to} \sum_{ (w, v) \in S}{w} < w_\mathrm{max}.
	\end{equation}
	
	For training, we generated $20$ training instances with $w_\mathrm{max} = 10$, with weights and values uniformly sampled from $[0, 1]$, and ten more instances for validation.

	\subsubsection{Traveling Salesman Problem}
	The traveling salesman problem is another type of combinatorial optimization problem. We consider the case of finding a Hamiltonian cycle of maximum weight within a fully connected, weighted, undirected graph. We formulate it as a maximization problem 
	(this is equivalent via a transformation of edge weights to the common formulation as a minimization problem).

	For training, we use $40$ different graphs with weights uniformly sampled from $[0, 1]$. For evaluation, another 10 problem graphs are used.
	
	\subsubsection{Continuous Function Optimization}\label{continuous-function-optimization}
	
	For continuous optimization, nineteen standard benchmark $\mathbb{R}^2 \mapsto \mathbb{R}$ objective functions, as defined in Ref.~\autocite{simulationlib}, are used. The goal in each case is to find the global minimum. The Ackley function, Beale function and Levy function \#13 are used for validation. The following functions are used for training: Rastrigin, Rosenbrock, Goldstein--Price, Bukin~\#6, Matyas, Cross-in-Tray, Eggholder, Holder, McCormick, Schaffer~\#2, Schaffer~\#4, Styblinski-Tang, Sphere, Himmelblau, Booth, Three-Hump Camel. While the Beale function is plateau-shaped, except for its steep borders, the Ackley function and the Levy function \#13 are highly rugged with a considerable number of local optima.

	For data normalization purposes, we rescale and translate the functions so that their domain is $[-1, 1] \times [-1, 1]$ and subtract their minimum value. Obviously, this normalization is only possible because the minimum value is already known. While this is not representative of real-world problems, it is still sufficient for investigating whether evolutionary algorithms with deep reinforcement learning can be applied in continuous problem domains at all.

	\subsection{Baseline Evolutionary Algorithms}\label{baseline-algorithms}
	To solve the three types of benchmark problems, we use baseline evolutionary algorithms, which we shall later enhance through deep reinforcement learning.

	The following paragraphs give a brief explanation of the specifics of these baseline algorithms, their configurable parameters, and how the fitness of individuals is defined.
	
	\subsubsection{Baseline Algorithm for the 0-1 Knapsack Problem}
	\label{baseline-knapsack}

	In the evolutionary algorithm used for the knapsack problem, solutions are encoded as binary vectors. Fitness is defined as the sum of weights of the selected items.
	
	The initial population is created by randomly generating binary vectors with equal probability for $0$ and $1$. To ensure that the weight limit is not exceeded, items are randomly removed from invalid candidate solutions until the constraint is fulfilled.
	
	Parent selection is performed through \emph{tournament selection} with tournament size $2$. In a tournament, two individuals are randomly taken from the population and the fitter one is selected as parent. Two tournaments are performed for each pair of parents. The winner of the first tournament does not participate in the second one, but can again be selected in any future pair of tournaments.
	
	Recombination is performed through uniform crossover. With a probability of $1 - \texttt{crossover\_rate}$ the parents are directly copied into the offspring generation. Else, two children are created by combining the parent genomes. For each gene
	(i.e.~entry of the binary vector), there is a $50\%$ chance of child $1$ inheriting from parent $1$ and child $2$ inhering from parent $2$. Else, child $1$ inherits from parent $2$ and child $2$ inherits from parent $1$.
	
	All children then undergo mutation. Each bit is flipped with a probability of $\texttt{mutation\_rate}$.
	
	Survivor selection is performed using an elitism mechanism, which ensures that the fitness of the best individual in a population never degrades. The $\texttt{elite\_size}$ fittest individuals from the parent population and the $\texttt{population\_size} - \texttt{elite\_size}$ fittest offspring individuals are selected for survival into the next generation.

	\subsubsection{Baseline Algorithm for the Traveling Salesman Problem}\label{baseline-tsp}

	In the evolutionary algorithm used for the traveling salesman problem, integer-valued genes are used.
	For a graph with $n$ nodes, a solution is encoded as a permutation $(a_0, a_1, \dots, a_{n-1})$ of $(0, 1, \dots, n-1)$. 
	The fitness of a solution is calculated as 
	\begin{equation}
	\sum\limits_{i=0}^{n-1} w_{a_i, a_{i+1\  \mathrm{mod}\ n}},
	\end{equation}
	where $w_{i, j}$ is the weight of the edge between nodes $i$ and $j$. The initial population is a set of random permutations.
	
	Like in the evolutionary algorithm for the knapsack problem, parents are chosen via tournament selection (see \cref{baseline-knapsack}). The different crossover operators (described below) only generate one child for each pair of parents, so twice the number of tournaments have to be performed for the same population size.
	
	We use the traveling salesman problem to evaluate the ability of a reinforcement learning agent to select from a set of different operators.
	To this end, we employ the following seven crossover operators:	\emph{one-point crossover}, \emph{two-point crossover}, \emph{linear-order crossover}, \emph{cycle crossover}, \emph{position-based crossover}, \emph{order-based crossover}, and \emph{partially mapped crossover}, as explained in Ref.~\autocite{Anand2016}.
	Depending on the crossover operator, children inherit sub-paths, the relative order of nodes, the position of nodes, or a combination thereof, from their parents. The probability of performing crossover, instead of directly copying the parents into the offspring population, is defined by the $\texttt{crossover\_rate}$ parameter.
	
	Mutation is performed through inversion, as follows. Each child is mutated with a probability defined by $\texttt{mutation\_rate}$. If the child is mutated, two positions in its genome are randomly chosen. The order of all the genes between these two positions is then inverted.
	
	Survivor selection is performed with the same elitism mechanism used for the 0-1 knapsack problem.

	\subsubsection{Baseline Algorithm for Continuous Function Minimization}
	\label{baseline-continuous}

	The evolutionary algorithm for minimization of $\mathbb{R}^2 \rightarrow \mathbb{R}$ functions represents candidate solutions as real-valued vectors. For self-adaptive mutation, each genome also encodes a positive, real-valued step-size $\upsilon$. The fitness of a solution $(x_1, x_2)$ evaluated on a function $g$ is defined as 
	\begin{equation}
	1 / \max(g(x_1, x_2), 10^{-20}).
	\label{continuous-objective-clipped}
	\end{equation}
	Taking the reciprocal value turns the minimization into a maximization problem, so that our definitions from previous sections are consistent across all problem classes. The $\max$-operator prevents problems with floating point calculations.
	
	The initial population is generated by uniformly sampling from the function domain. The step-size of each individual is first set to $\texttt{initial\_step\_size}$.
	
	Evolutionary pressure is induced by only selecting the $\texttt{parent\_percentage} \times \texttt{population\_size}$ fittest individuals as a set of parents for mutation. No crossover operator is used.
	
	For each offspring individual, a parent from the parent set is randomly selected and then mutated through one-step self-adaptive mutation, as follows:
	First, the step-size $\upsilon_i$ of individual $i$ is multiplied with $\max(\upsilon_i, \texttt{min\_step\_size})$, where $\upsilon_i$ a sample from the log-normal distribution $e ^ {\mathcal{N}(0, \tau)}$, with self-adaptation strategy parameter $\tau$.
	Then, a sample from $\mathcal{N}(0, \upsilon_i)$ is taken for each gene and added onto the current value.
	If mutation leads an individual to leave the function's domain, it is re-initialized at a uniformly sampled random coordinate, and $\upsilon$ is reset to $\texttt{initial\_step\_size}$.
	
	The same elitism mechanism as in the other baseline algorithms is used for survivor selection.

	\subsection{Adaptation Methods}\label{subsection-adaptation-methods}
	Now that we have established the baseline algorithms, we propose different ways of enhancing them through reinforcement learning.
	Each of the proposed adaptation methods replaces or enhances one component of the evolutionary algorithm (parent selection, crossover, mutation, or survivor selection, as explained in \cref{evolutionary-computation}). To show the range of possibilities for applying deep reinforcement learning to evolutionary algorithms, we propose methods for all levels of adaptation explained in Section \ref{levels-of-adaptation}).
	
	Recall that our reinforcement learning algorithm learns a stochastic policy (see \cref{reinforcement-learning-methods}), meaning that actions are taken by sampling from a probability distribution conditioned on the neural network's parameters $\theta$ and its input. We use the following probability distributions, which have different definition domains and are therefore useful for taking different types of actions:
	\begin{itemize}
		\item Bernoulli trials are used for discrete binary actions, as sampling from them returns either $0$ or $1$. The neural networks outputs a probability $p_\theta \in [0, 1]$ to parameterize the distribution. \arxiv{
			We use Bernoulli trials to:
			\begin{itemize}
				\item select subsets of the population as parents (Section~\ref{parent-selection}),
				\item decide which bits should be mutated in the evolutionary algorithm with binary encoding. (Section~\ref{mutation-chromosome-binary})
			\end{itemize}
		}
		\item Beta distributions can be used for real-valued, constrained policies, as proposed in Ref.~\autocite{Chou17}. Sampling from them yields a number between $0$ and $1$. For a unimodal beta distribution, the neural network has to output two scalars $\alpha_\theta, \beta_\theta \in (1, \infty)$. \arxiv{We use beta distributions to:
			\begin{itemize}
				\item control the mutation rate (explained in Section~\ref{baseline-knapsack}) for the entire population (Section~\ref{mutation-rate-control-global}),
				\item control the mutation rate of each individual separately (Section~\ref{individual-mutation-rates}).
		\end{itemize}}
		
		\item Categorical distributions are useful for selecting a single action from a finite set of $k$ discrete actions. The distribution is parametrized by probabilities ${(p_\theta)}_i$ for each element~$i$ to be selected. \arxiv{We use a categorical distribution to:
			\begin{itemize}
				\item select from a set of different crossover operators (listed in Section~\ref{baseline-tsp}) on the fly (Section~\ref{operator-selection}).
			\end{itemize}
		}
		
		\item Normal distributions are utilized for real-valued, unbounded actions. Normal distributions are parameterized by a mean $\mu_\theta \in \mathbb{R}$ and a standard deviation $\sigma_\theta \in \mathbb{R}_+$. \arxiv{We use normal distributions to:
			\begin{itemize}
				\item alter (multiplicatively) the fitness of individuals to influence the selection of parents (Section~\ref{fitness-shaping-parent}),
				\item alter (overwrite) the strategy parameter $\tau$ (explained in Section~\ref{baseline-continuous}) for the entire population (Section~\ref{strategy-parameter-control-global}),
				\item alter (overwrite) the strategy parameter $\tau$ separately for each individual (Section~\ref{individual-strategy-parameters}),
				\item alter  (overwrite) the step-size $\upsilon$ (explained in Section~\ref{baseline-continuous}) of each individual in the population (Section~\ref{methods-individual-step-sizes}),
				\item alter (overwrite) step-sizes for each gene of each individual (Section~\ref{mutation-chromosome-continuous})
				\item alter (overwrite) the fitness value of individuals to influence survivor selection, in order to select a fixed number of survivors (Section~\ref{survivor-selection}).
			\end{itemize}
		}
	\end{itemize}

	In some cases, we apply a function ($\exp$ or $\mathrm{softplus})$ to the samples from a normal distribution. Note that in the context of the proximal policy optimization algorithm, we treat the sample from the normal distribution as the action. The subsequent transformations are part of executing the action and are not considered in the gradient calculation.
	
	The following sections explain the details of how these probability distributions are used by the different adaptation methods, on an implementation-independent level. \arxiv{In \cref{network-architecture} we then define how the neural network that controls the distributions operates, how its inputs are encoded and how the constraints on its output domains are enforced.}
	
	Note that sampling from a random distribution parameterized by the neural network means that the network uses information about the current situation (see Section~\ref{network-architecture}), i.e.~the randomness is intelligently constrained rather than arbitrary.

	\subsubsection{Environment-Level Adaptation}\label{environment-level}
	On the environment level, we let an agent alter or replace the fitness function without using handcrafted auxiliary functions. Altering the fitness landscape could allow for more diverse populations, which could help in exploring more of the solution space.
	
	\paragraph{Fitness Shaping}\label{fitness-shaping-parent}
	In fitness shaping, we sample a vector $\varepsilon \in \mathbb{R}^{\texttt{population\_size}}$ from a set of $\texttt{population\_size}$ normal distributions parameterized by the neural network, and multiply it elementwise with the population's fitness values, before applying the parent selection mechanism of the baseline algorithm. On the continuous problem set, we multiply fitness values with $\exp(\varepsilon)$, as the difference in fitness values is typically much larger.
	
	\paragraph{Survivor Selection}\label{survivor-selection}
	In survivor selection, we assign each individual from the parent and offspring population a fitness value by sampling from $2 \cdot \texttt{population\_size}$ independent normal distributions parameterized by the neural network. We then select the $\texttt{population\_size}$ individuals with the highest fitness value for survival. Unlike in fitness shaping, the learned fitness function does not merely alter the objective function, but replaces it entirely.
	
	\subsubsection{Population-Level Adaptation}
	On the population level, we propose two methods that dynamically control the mutation rate / strategy parameter of the baseline evolutionary algorithms. This could -- for example -- enable a coarse-to-fine approach to optimization, in which the amount of mutation decreases over time.  We also propose a method for selecting from the set of crossover operators for the traveling salesman problem, which could allow the evolutionary algorithm to explore along better trajectories in the solution space, as different operators let children inherit different features from their parents. These methods work as follows:
	
	\paragraph{Mutation Rate Control}\label{mutation-rate-control-global}
	To control the mutation rate on the population level, we sample a value $\in [0, 1]$ from a single beta distribution parameterized by the neural network.
	
	\paragraph{Strategy Parameter Control}\label{strategy-parameter-control-global}
	To control the strategy parameter of the evolutionary algorithm for continuous optimization on the population level, we sample a value $\tau' \in \mathbb{R}$ from a single normal distribution parameterized by the neural network, and use the $\mathrm{softplus}$ nonlinearity  to calculate the positive-valued strategy parameter $\tau = \mathrm{softplus}(\tau') = \log(1 + e^{\tau'})$.
	
	\paragraph{Operator Selection}\label{operator-selection}
	To select from the set of available crossover operators for the traveling salesman problem, we sample from a categorical distribution parameterized by the neural network \arxiv{ (where each category corresponds to a crossover operator)}.
	
	\subsubsection{Individual-Level Adaptation}
	The first two methods for individual-level adaptation use the same continuous mutation parameters as on the population level, but control them separately for each individual. Next, we propose an alternative way of controlling self-adaptation in evolution strategies. Controlling mutation per individual could increase the capability of the evolutionary algorithm to deal with diverse populations, for example by mutating low-fitness individuals more.  Finally, we introduce a way of letting a learning agent directly perform the parent selection processes of an evolutionary algorithm. This could allow us to guide the population through the fitness landscape more deliberately than the baseline methods do. These methods work as follows:
	
	\paragraph{Mutation Rate Control}\label{individual-mutation-rates}
	To control the mutation rate per individual, we sample from  $\texttt{population\_size}$ independent beta distributions parameterized by the neural network.
	
	\paragraph{Strategy Parameter Control}\label{individual-strategy-parameters}
	To control the strategy parameter per individual, we sample values $(\tau_1', \dots, \tau_\texttt{population\_size}')$ from a set of independent normal distributions parameterized by the neural network, and then calculate the strategy parameter for individual $i$ as $\tau_i = \mathrm{softplus}( \tau_i' )$.
	
	\paragraph{Step-Size Control}\label{methods-individual-step-sizes}

	Instead of controlling strategy parameters to indirectly influence the evolution of step-sizes, 
	
	the step-size control method lets the neural network output multipliers for the step-sizes more directly. To do so, the step-size $\upsilon_i$ of individual $i$ is changed multiplicatively via $\upsilon_i \gets \mathrm{softplus}(\xi_i) \upsilon_i$, where $\xi_i$ is a sample from a normal distribution parameterized by the neural network. 
	The step-sizes are then used to mutate the genes of the individuals, as in the baseline algorithm.
	
	\paragraph{Parent Selection}\label{parent-selection}
	To select a subset of parents, we sample a binary vector $x \in \{0, 1\}^\texttt{population\_size}$ from a set of independent Bernoulli distributions parameterized by the neural network. $\mathrm{Iff}$ $x_i=1$, individual $i$ becomes a parent candidate for the offspring population.In the evolutionary algorithm for the knapsack problem, we let the agent perform a pre-selection of parent candidates, and then apply the baseline parent selection method to create pairings. In the evolutionary algorithm for continuous optimization there is no parent-pairing step, so this adaptation method directly controls which parents produce offspring.
	
	\subsubsection{Component-Level Adaptation}
	The last class of proposed adaptation methods is component-level adaptation. We propose a method for mutating binary genes and a method for mutating real-valued genes. Component-level mutation could allow the agent to more directly control the direction in which individuals move through the solution space.
	
	\paragraph{Binary Mutation}\label{mutation-chromosome-binary}
	To directly control binary mutation, we sample a matrix $\in \{0, 1\}^{\texttt{population\_size} \times \texttt{genome\_size}}$ from independent Bernoulli distributions parameterized by the neural network. Each element corresponds to a gene in one specific individual of the population. If an entry of the matrix is $1$, the gene value is inverted.
	
	\paragraph{Step-Size Control}\label{mutation-chromosome-continuous}
	For component-level adaptation in the evolutionary algorithm for continuous optimization, we assign each individual $i$ a vector $(\upsilon_{i,1}, \dots, \upsilon_{i, \texttt{genome\_size}})$ of step-sizes. These step-sizes are multiplicatively mutated via $\upsilon_{i,j} \gets \mathrm{softplus}(\xi_{i,j}) \upsilon$, where $\xi_{i,j}$ is a sample from a normal distribution parameterized by the neural network. Each solution-encoding gene $k$ of individual $i$ is then mutated by adding a value sampled from $\mathcal{N}(0, \upsilon_{i, k})$, similarly to the baseline algorithm. 
	Through this mechanism, offspring is sampled from a multivariate Gaussian distribution with a diagonal covariance matrix. Alternatively, this can be interpreted as a trainable diagonal preconditioner, learning to rescale the fitness landscape around each parent individual to facilitate optimization. This could allow the evolutionary algorithm to make more deliberate decisions regarding the direction of mutation, compared to using the same step size along all problem dimensions or altering step sizes through a random process with static parameters (as in the baseline algorithm).

	\subsection{Network Architecture}\label{network-architecture}
\begin{figure}[]
		\centering
		\includegraphics[width=0.95\linewidth]{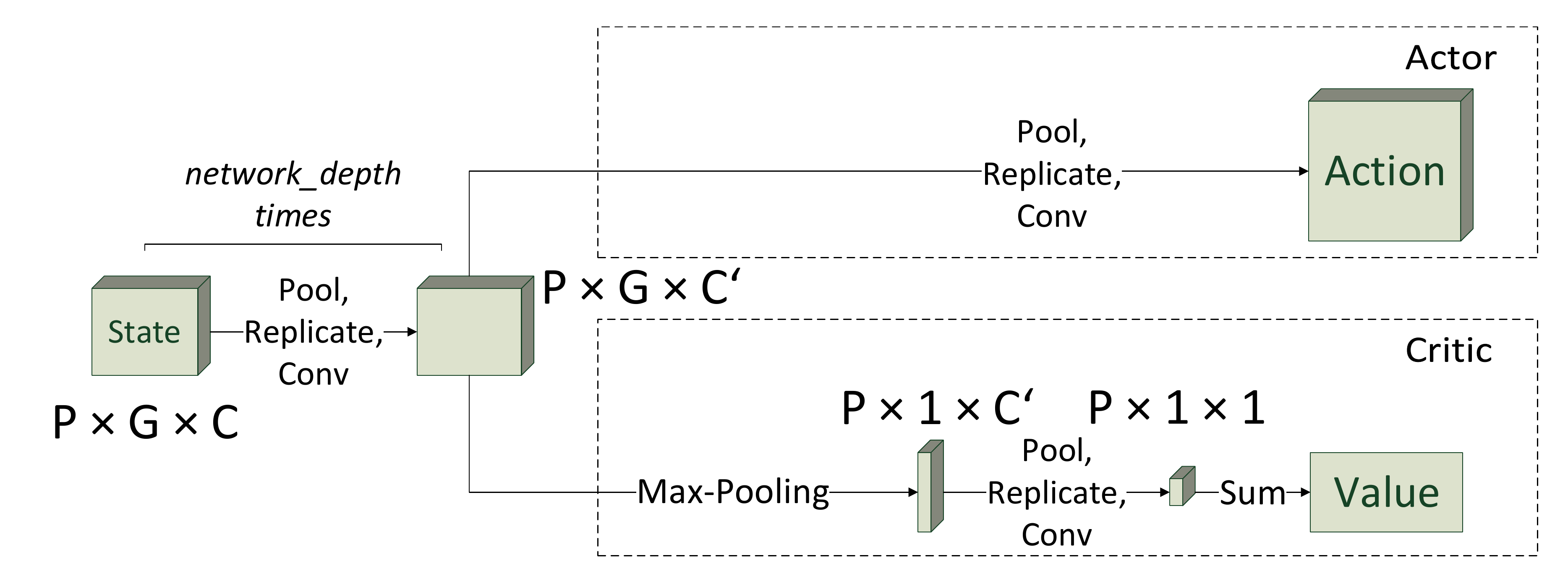}
		\caption{The overall neural network architecture. $P$, $G$ and $C$ are the population size, genome size and the number of channels, respectively. The dimensionality of the output action can be further reduced through max-pooling, depending on the adaptation method. Actor and critic operate on the same low-level features extracted by the
			''Pool, Replicate, Conv'' substructure visualized in~\cref{fig-network-architecture-prc}.}
		\label{fig-network-architecture}
	\end{figure}

\begin{figure}[]\label{}
		\centering
		\includegraphics[width=0.95\linewidth]{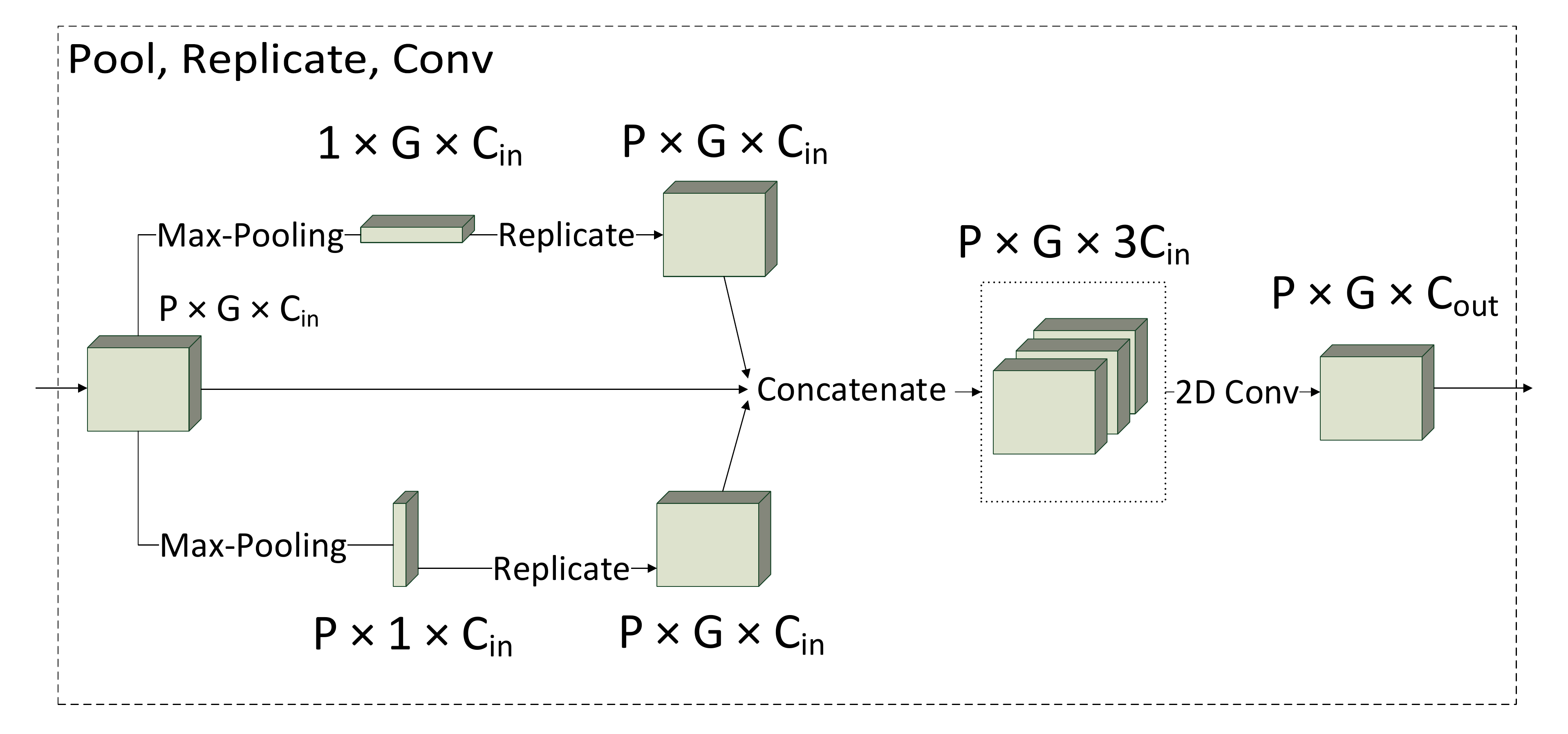}
		\caption{The "Pool, Replicate, Conv"-substructure of the neural network. $P$ and $G$ are the population and genome size, respectively. $C_\texttt{in}$ and $C_\texttt{out}$ are the numbers of input and output channels, respectively. Global features are extracted by pooling along either of the two ``spatial'' dimensions. They are then replicated along the same dimension, combined with local features through concatenation, and processed by a 2D convolutional layer with kernel size $1 \times 1$. This architecture ``broadcasts'' parts of global information to each individual and each gene, and is equivariant under permutations of individuals and of genes, i.e.~treats them equally.}
		\label{fig-network-architecture-prc}
	\end{figure}

	To perform the calculations for our adaptation methods, we propose the use of a 2D convolutional neural network (see~\cref{fig-network-architecture,fig-network-architecture-prc}). This section describes the requirements that a neural network architecture should fulfill in our application as well as a specific network architecture that fulfills these requirements.
	
	\subsubsection{Requirements}
	Instead of relying on hand-crafted features, the neural network should be offered as much information as possible about the state of the evolutionary algorithm and the problem instance, so that it can then extract the relevant features itself.
	
	The order in which individuals are stored in the computer should not affect the results. More specifically, there is no fixed index~$i$ reserved for individuals that across all problem instances have a specific special role; in other words, the meaning of the order of individuals is not persistent across problem instances. Hence, treatment of individuals should depend only on their features and not their order; in other words, the operations performed by the network should be equivariant under permutation of this order.
	
	Since \emph{in the case of our problem classes} the same holds for the order of genes, the network should also be equivariant under permutation of the gene order.
	
	A learned adaptation strategy should be useable with varying values of $\texttt{population\_size}$ and (in the case of our problem classes, where no chromosome index has a special role across all problem instances) varying values of $\texttt{genome\_size}$.
	
	Information about the entire population might be relevant for taking good actions. Hence, the network should extract and use features of the entire population. These features should be permutation-invariant, for the same reasons as the permutation-equivariance explained above. Similarly, information about all genes might be relevant, so that the network should also extract and use features of entire genomes (permutation-invariant features \emph{in the case of our problem classes}; see above).
	
	As described in~Section~\ref{actor-critic}, there are likely features that are relevant to both the actor and critic element. To avoid redundancy and facilitate training, the actor and critic should operate on the same low-level features.

	We use different types of probability distributions for the different adaptation methods introduced in~\cref{subsection-adaptation-methods}. Each distribution type has different parameters and therefore requires different output nonlinearities. For example, a categorical distribution requires probabilities in $[0, 1]$.
	
	Depending on the level of adaptation that a method operates on, the neural network has to output the parameters of a probability distribution for each gene, each individual or for the entire population. The dimensionality of the neural network's output is chosen accordingly.

	\subsubsection{Realization}
	The following input representation and architecture fulfill the aforementioned requirements:
	
	The network input is defined as a 4D array of size $\texttt{population\_size} \times \texttt{genome\_size} \times \texttt{num\_channels}$ (and a \texttt{batch\_size} dimension). Population-wide information (e.g.~the number of remaining generations) is replicated across the two trailing dimensions. Information about individuals (e.g.~their fitness) is replicated along the \texttt{genome\_size} dimension. We use the following feature channels for the knapsack problem and continuous optimization:
	\begin{itemize}
		\item Knapsack problem: Individual genomes, fitness values, the remaining number of generations, the weight limit, the weight of each item, and the value of each item.
		\item Continuous optimization: Individual genomes, logarithm of fitness values, the remaining number of generations, and individual step-sizes.
	\end{itemize}
	On the traveling salesman problem, we control the selection of crossover operators. Therefore, the input is based on pairs of parent individuals. Parents of a pair are assigned an arbitrary order. The input channels for a traveling salesman problem instance with $N$ nodes are:
	
	\begin{itemize}
		\item Individual genomes of first parents,

		individual genomes of second parents, fitness of first parents, fitness of second parents, the remaining number of generations, $N$ distance information channels for first parents, $N$ distance information channels for second parents.
		In each block of $N$ distance information channels, entry $(i, j)$ of channel $k$ contains the distance from node $g_{i,j}$ to node $k$, with $g_{i,j}$ being the the value of gene $j$ (i.e.~the $j^{\textrm{th}}$ visited node) of individual $i$.
		We plan a different representation for a future version of this work.
	\end{itemize}

	We then extract shared hidden features for the actor and critic through 2D convolutional layers with kernel size $1 \times 1$. To propagate global information, we perform the following operation before each convolutional layer separately along the $\texttt{population\_size}$ and $\texttt{genome\_size}$ dimension: For each channel, the maxima along the respective dimension are calculated. The resulting vector is then replicated along the same dimension, yielding a new matrix of size $\texttt{population\_size} \times \texttt{genome\_size}$.
	The global features extracted through successive pooling and replication can then be processed together with local features by the next convolution filter. This process of pooling, replication and convolution is illustrated in~\cref{fig-network-architecture-prc}.

	For the critic's output, we eliminate the \texttt{genome\_size} dimension through max-pooling. We then add population-wise features through max-pooling and replication along the \texttt{population\_size} dimension and apply one more convolutional layer with one $1 \times 1$ filter. The resulting scalars are summed up to calculate the value estimate.
	
	For the actor, we simply apply one more step of global pooling, replication and convolution to the shared hidden features. This is followed by max-pooling along the \texttt{genome\_size} dimension or both the \texttt{genome\_size} and \texttt{population\_size} dimension, if a vector or scalar output is required.
	To fulfill the constraints on the output domain for the different adaptation methods, we use the following output nonlinearities
	\begin{itemize}
		\item Bernoulli distribution: A single channel for $p \in [0, 1]$ with the nonlinearity ${\mathrm{sigmoid}(z) = \frac{1}{1 + e^{-z}}}$.
		\item Normal distribution: One channel for $\mu \in \mathbb{R}$ without any nonlinearity. One channel for $\sigma \in \mathbb{R}_{+}$ with the nonlinearity $\mathrm{softplus}(z) = \ln(1+e^z)$.
		\item Beta distribution: Two channels for $\alpha, \beta \in [1, \infty)$, with the nonlinearity $\mathrm{softplus}(z) + 1$.
		\item Categorical distribution: One channel for each of the $k$ category probabilities $p_i \in [0, 1] : \sum_{i=0}^{k-1}{p_i} = 1$, using the $\mathrm{softmax}$ nonlinarity 
		\begin{equation}
		p_i = \frac{e^{z_i}}{\sum\limits_{j=0}^{k-1} e^{z_j}},
		\end{equation}
		where $(z_0, \dots , z_{k-1})$ are the network activations before the nonlinearity.
	\end{itemize}

	\subsection{Evaluation Methods}\label{experimental-setup}
	To evaluate the usefulness of the different proposed adaptation methods, they are compared to the baseline algorithms. To do so, we  use the performance metrics and the evaluation procedure defined in the following sections.
	
	\subsubsection{Performance Metrics}\label{metrics}
	We use two performance metrics to evaluate our evolutionary algorithms: mean best fitness ($\mathrm{MBF}$) and mean best function value ($\mathrm{MBFv}$):
	
	\arxiv{
		\begin{itemize}
			\item Mean best fitness is the fitness value of the fittest individual in the population, per generation, averaged over multiple runs of the evolutionary algorithm.
			\item Mean best function value is the lowest objective function value found by an individual in the population, per generation, averaged over multiple runs of the evolutionary algorithm.
		\end{itemize}
		We use MBF to assess the performance in combinatorial optimization and MBFv to assess the performance in continuous optimization. We use the mean best function value because we want to assess the quality w.r.t.\ to the objective function, not the clipped fitness function from \cref{continuous-objective-clipped}.
	}
	
	We refer to the average fitness / function value achieved in the final episode as terminal mean best fitness ($\mathrm{tMBF}$) / terminal mean best function value ($\mathrm{tMBFv}$).
	
	\subsubsection{Evaluation Procedure}\label{procedure}
	Each experiment is about optimizing one evolution parameter for one of the three problem classes (knapsack, traveling salesman, continuous optimization). A fine-grained search for an optimal (but \emph{static}) value of that evolution parameter within the baseline algorithm is compared to our methods that learn to (\emph{dynamically}) control that evolution parameter.
	
	During each experiment on one evolution parameter, all other evolution parameters are held fixed at their default values. These default values are determined in advance by a coarse grid-search with the baseline algorithms. The coarseness of the search allowed a reasonable runtime (about 2 days). 
	The coarseness of the search also means that the thereby determined default evolution parameters are not perfect. However, this is okay, because our algorithms and the baseline algorithms work with the same set of fixed values for the evolution parameters (except the parameter on which \emph{static} vs.\ \emph{dynamic} fine-tuning is compared). We deliberately chose to set the default elite size to $0$ (i.e.~the entire population is replaced in each generation), as we found this to magnify the impact of the remaining parameters, allowing for a better assessment of the quality of different adaptation methods. The default evolution parameter values are listed in~\cref{ea-baseline-default-params}. The fine-tuned evolution parameter values are listed in~\cref{ea-baseline-optimised-params}.
	
	When fine-tuning the discrete parameters (elite size, number of parents, crossover operators), we tested all possible values.
	
	For the mutation rate parameter (values in $[0,1]$, we searched the best-performing range of parameters $[0.005, 0.013]$ with a step size of $0.0001$.
	For the mutation parameter for continuous optimization (values in $\mathbb{R}_{+}$), we searched in the best-performing range $[0, 1]$ with an accuracy of $2$ decimal digits. This accuracy appears sufficient, as we observed little to no difference in MBF(v) around the discovered optima.

	For each adaptation method, we use a separate training set to train $21$ agents using the same evolution parameters, which allows us to assess how reliably good policies can be learned. 
	Each agent is trained for $500$ iterations.

	The deep learning hyperparameter values we use are summarized in~\cref{deep-learning-hyper_params}:
	
	After training, we evaluate the mean best fitness / function value achieved by each agent on a separate validation set and compare it to that of the baseline algorithm.
	
	Mean best fitness is calculated over $100$ runs of the evolutionary algorithm. Mean best function value is calculated over $500$ runs of the evolutionary algorithm.
	
	When using beta or normal distributions, actions are not taken by random sampling during validation. Instead, the mean of the distribution is taken deterministically. We found that this improves performance after the limited number of training iterations, as one does not have to wait for the loss function to decrease the distribution entropy after convergence of the policy and value estimate (see Section~\ref{entropy-exploration}).

	\section{Results and Discussion}\label{results}

	\begin{figure}[]
		\centering
		
		\includegraphics[width=0.95\linewidth]{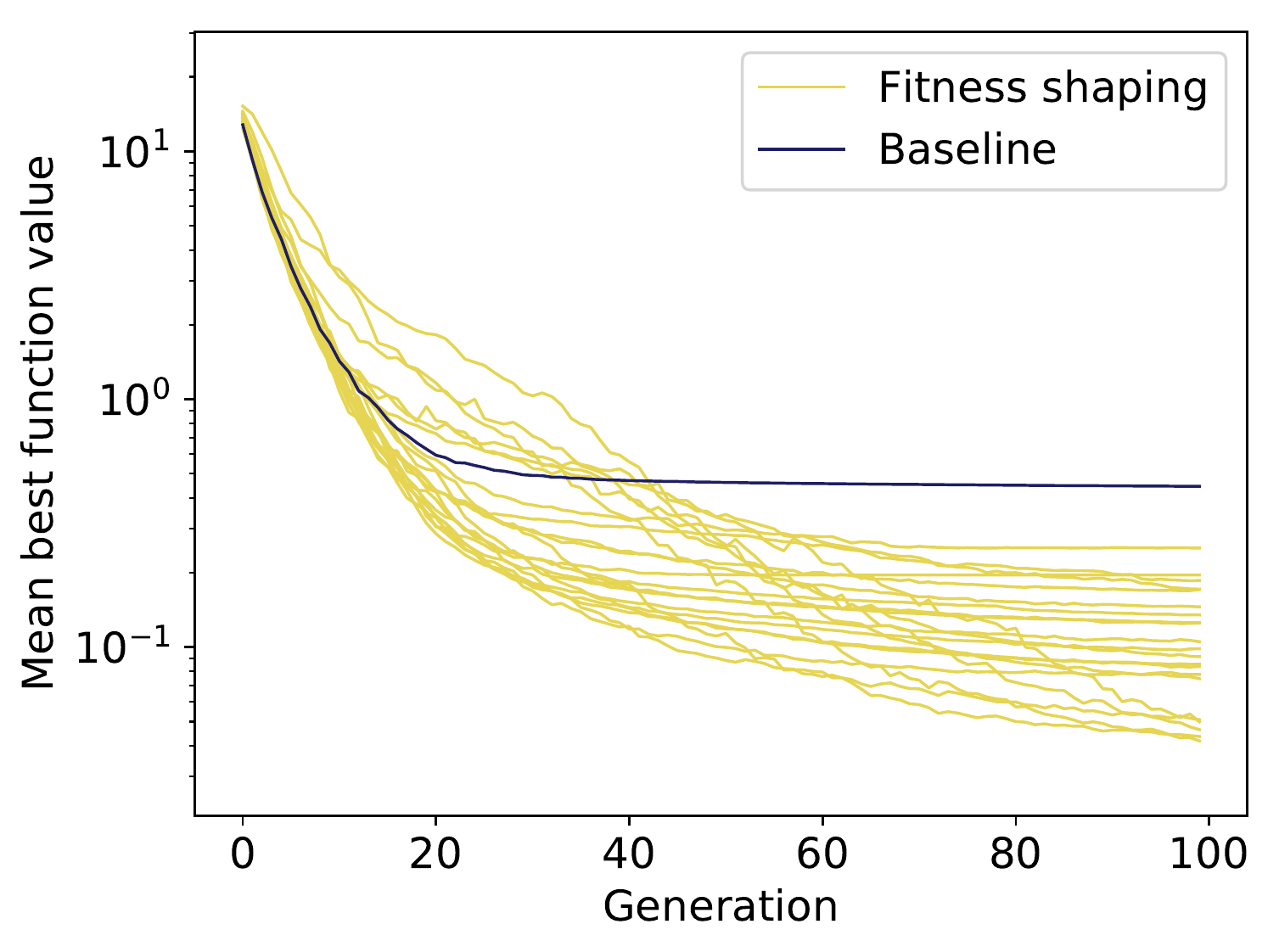}
		
		\caption{MBFv (smaller is better) of 21 agents trained for fitness shaping, evaluated on the Levy \#13 function, compared to the baseline algorithm.
			All trained agents achieved better-than-baseline performance up to a factor of $12$. This shows that \emph{learning to evolve} improves the results evolutionary algorithms. After a single training run, the user can expect above-baseline performance, but choosing the best out of multiple agents is likely to yield even better results.}
		
		\label{fig-results-environment-fitness_shaping-cont-2}
	\end{figure}
	
	\begin{figure}[]
		\centering
		\includegraphics[width=0.95\linewidth]{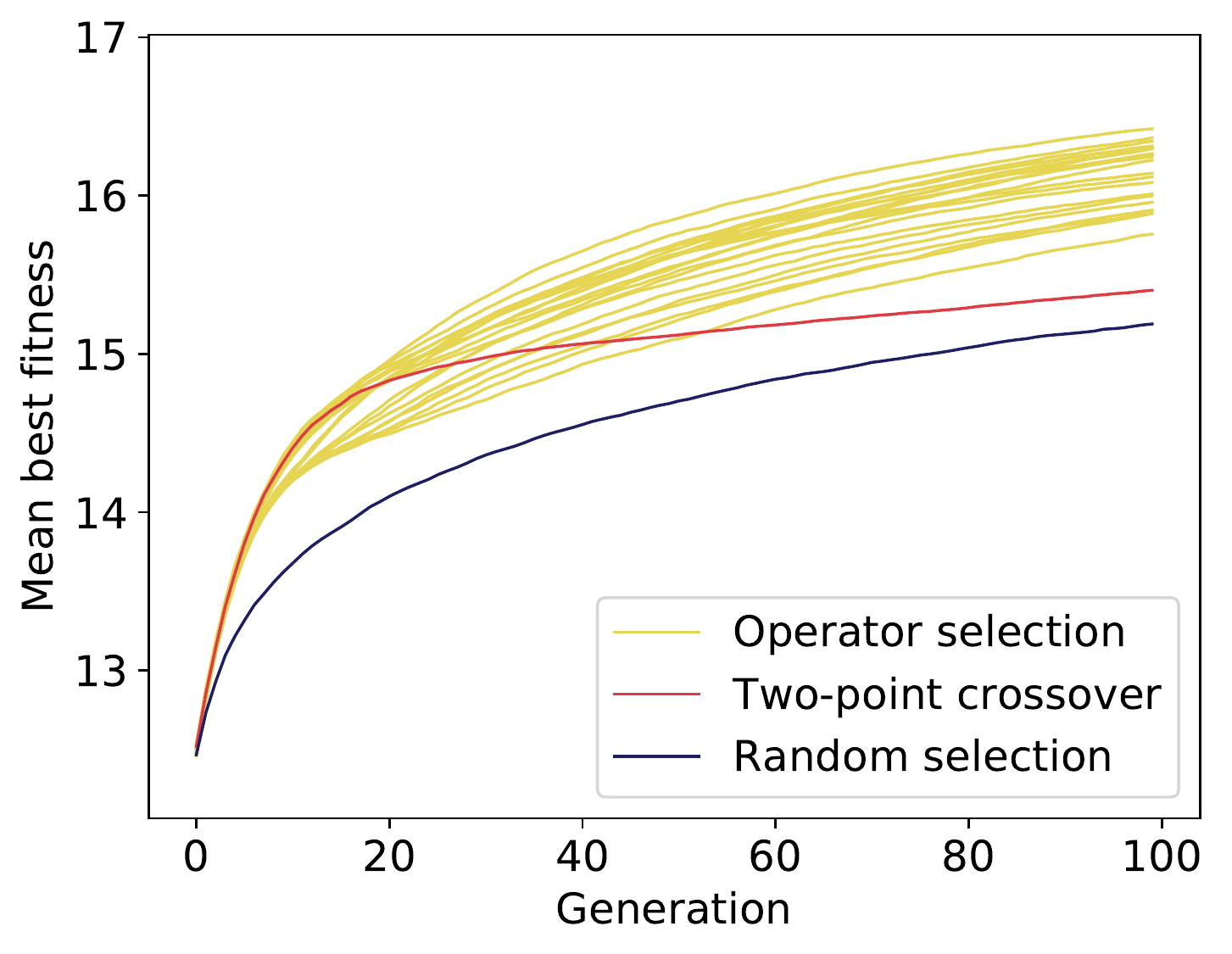}
		\caption{MBF (larger is better) of 21 agents trained for crossover operator selection on the traveling salesman problem, compared to the best-performing single operator, as well as random operator selection with uniform probability. All trained agents performed better than baseline from generation $55$ onward. The user can thus expect better-than-baseline performance even after training  only one agent. tMBF varied between $16.422$ and $15.757$, i.e.~selecting the best out of multiple trained agents is likely to yield even better results.}
		\label{fig-results-population-operator_selection}
	\end{figure}
	
	\begin{figure}[]
		\centering
		{\includegraphics[width=0.95\linewidth]{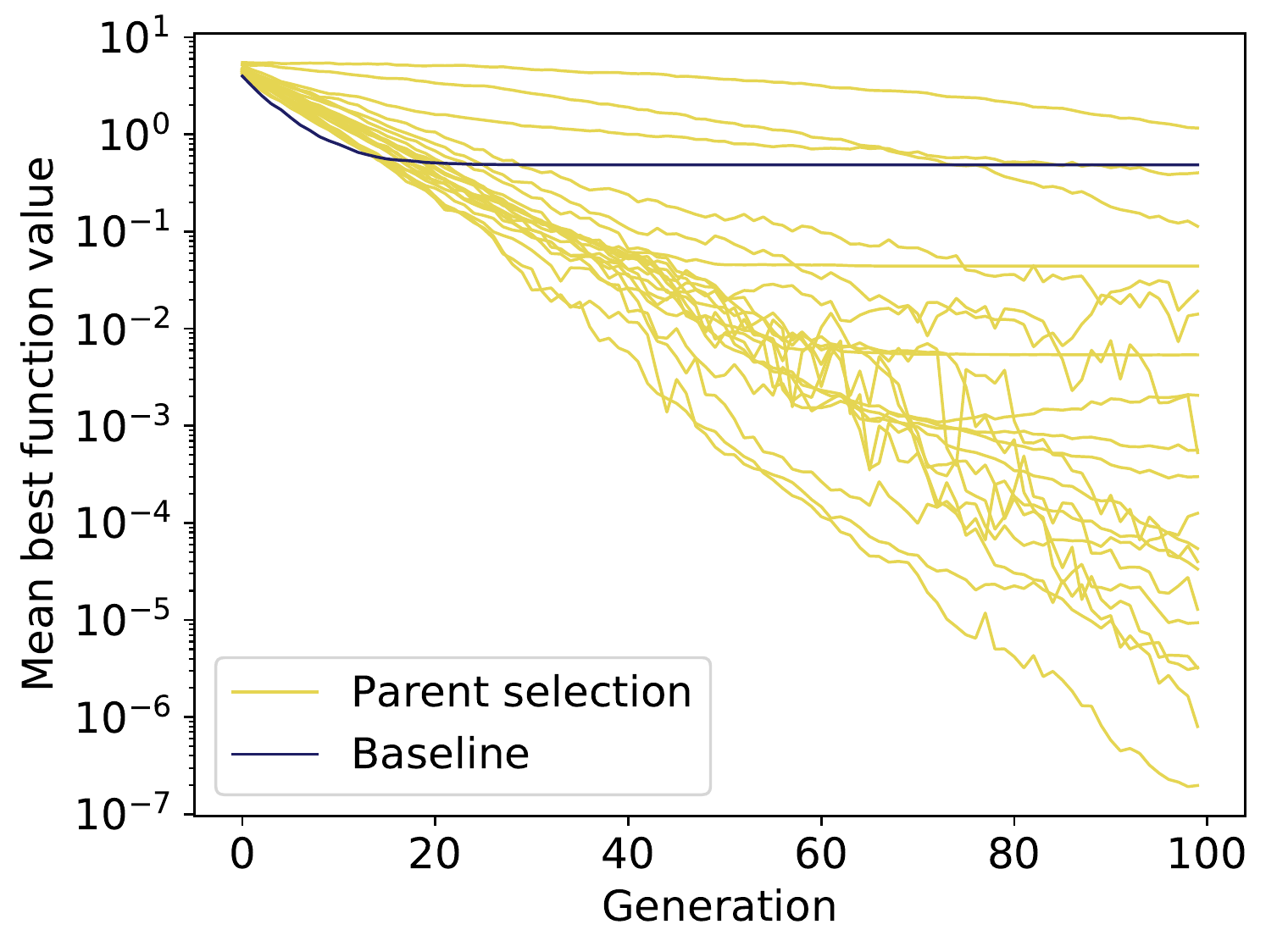}
			
			\caption{MBFv (smaller is better) of 21 agents trained for parent selection, evaluated on the Ackley function, compared to the baseline algorithm. All but one of the trained agents outperform the baseline algorithm by factors of up to $2.5 \cdot 10^6$ but tMBFv varies by multiple orders of magnitude among agents. The best agent exhibits a nearly exponential improvement in fitness across all generations. After a single training run, the user can expect above-baseline performance, but choosing the best out of multiple agents is likely to yield even better results.}
			\label{fig-results-individual-parent_selection-cont-0}}
	\end{figure}
	
	\begin{figure}[]
		\includegraphics[width=0.95\linewidth]{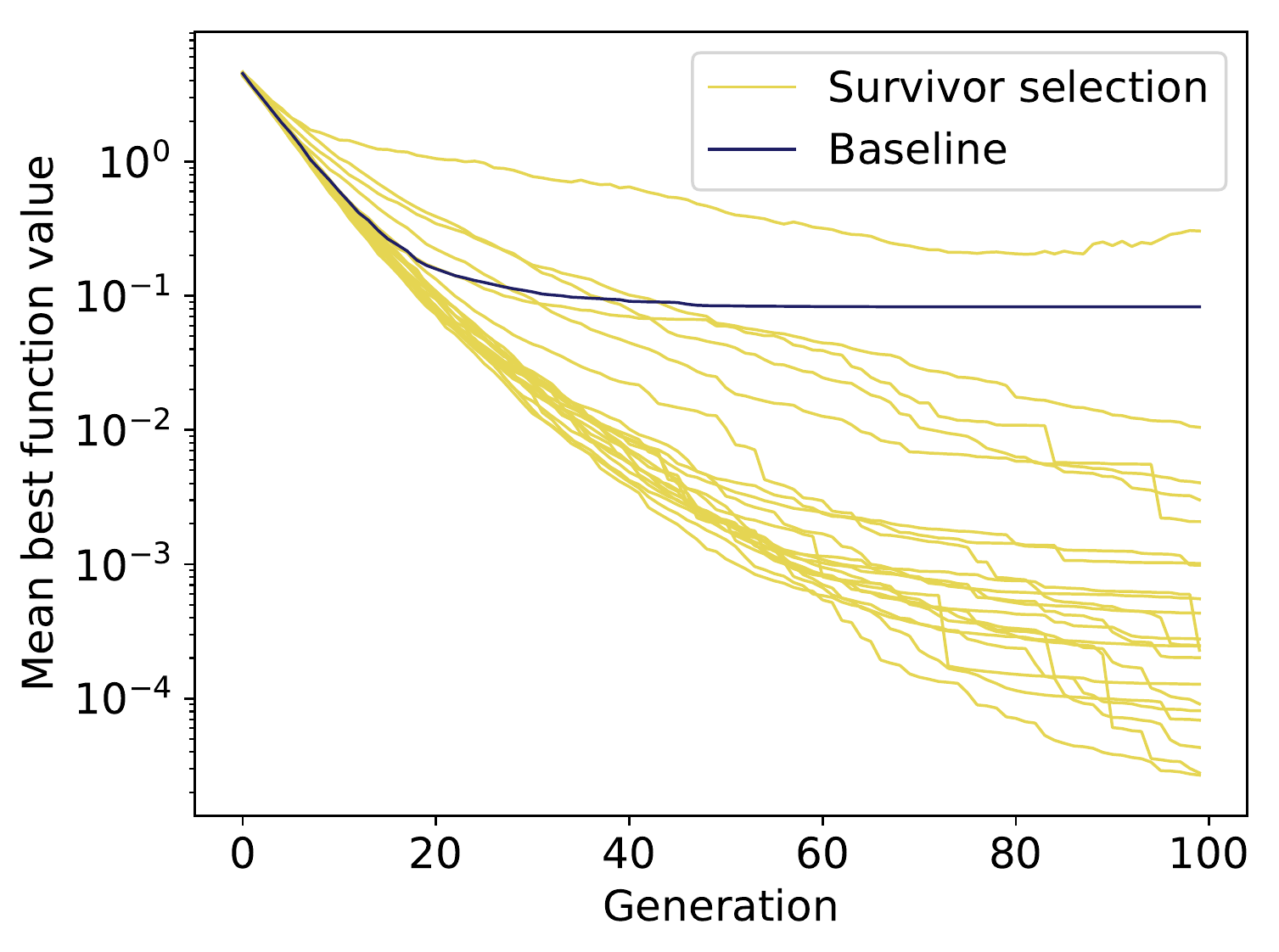}

		\caption{
		MBFv (smaller is better) of 21 agents trained for survivor selection, evaluated on the Ackley function, compared to the baseline algorithm. The baseline algorithm was outperformed by up to four orders of magnitude, but the variance in tMBFv values among agents was large. After a single training run, the user can expect above-baseline performance, but choosing the best out of multiple agents is likely to yield even better results.}
		\label{fig-results-individual-survivor_selection-cont-0}
	\end{figure}
	
	Following the evaluation procedure defined in \cref{procedure}, we first tuned the evolution parameters of the baseline evolutionary algorithms, before benchmarking our proposed adaptation methods against them. 
	
	In general, we found that agents could learn behavior with properties that compare favorably to the baseline algorithms: Achieving a better MBF(v) in fewer generations, not stagnating in fitness prematurely, or at least matching the performance of hand-crafted heuristics. We were successful in training agents both for discrete (e.g.~parent selection) and continuous (e.g.~mutation probability) action spaces, as well as for discrete and continuous optimization problems.
	Furthermore, methods from all different levels of adaptation were able to outperform the baseline algorithms.
	
	However, the adaptation methods differed considerably in their performance. In some cases, there were also large performance differences between agents belonging to the same method.
	We can distinguish between the following four cases, which relate to the suitability of the adaptation methods for the two considered use cases: Training with limited time/resources and training with much time/resources (see the beginning of \cref{methods}):
	
	\begin{enumerate}[wide, labelwidth=!, labelindent=1em, label=Case \arabic*)]
		\item All agents achieve similar or better performance than the baseline algorithm and the variance among agents is small. This is favorable for the use case with limited training time/resources, as one can expect to achieve good performance after training a single agent. The following adaptation methods belong to this case:
		
		\begin{itemize}
			\item Population-level mutation rate control (\cref{fig-results-population-mutation-knap}):
			All trained agents matched the performance of the baseline algorithm with an optimized mutation rate. This is remarkable, as the mutation rate only yields good results for a small range of parameter values (around 2\% of the valid interval $[0,1]$), as determined experimentally). Despite this difficulty, our reinforcement learning algorithm was able to learn a well-performing policy.
			
			\item Survivor selection, knapsack problem (\cref{fig-results-individual-survivor_selection-knap}): 
			While all agents ended with slightly below-baseline tMBF values (average of $15.81$, compared to $15.86$ of the baseline algorithm), they exhibited slightly higher MBF values during the first $40$ generations. Most importantly, they consistently learned a meaningful survivor selection mechanism that performed much better than replacing the population in each generation (tMBF of $15.43$). 
		\end{itemize}
		
		\item (Nearly) all trained agents achieve similar or better performance than the baseline algorithm, but the variance in performance among well-performing agents is large. One can expect to achieve good performance after training a single agent. But if more time/resources are available for training, selecting the best-performing agent out of several trained agents is likely to lead to even better results. The following adaptation methods pertain to this case:
		
		\begin{itemize}
			\item Fitness shaping, knapsack problem set (\cref{fig-results-environment-fitness_shaping-knap}):
			Most of the $21$ trained agents matched the performance of the baseline algorithm, but two out of $21$ trained agents achieved noticeably higher tMBF values.
			
			\item Fitness shaping, continuous problem set:
			Nearly all agents outperformed the baseline algorithm. The best agents were better by factors of up to approximately $10^3$, $2$, and $10$ on the Ackley (\cref{fig-results-environment-fitness_shaping-cont-0}), Beale (\cref{fig-results-environment-fitness_shaping-cont-1}) and Levy \#13 (\cref{fig-results-environment-fitness_shaping-cont-2}) function, respectively.
			
			\item Operator selection (\cref{fig-results-population-operator_selection}):
			From generation $55$ onward, all trained agents achieved higher MBF values than the deterministic application of the best crossover operator (two-point crossover) and than random operator selection with uniform probability.
			
			\item Parent selection, knapsack problem set (\cref{fig-results-individual-parent_selection-knap}):
			Except for one outlier, all agents reached baseline or above-baseline performance, with the highest tMBF being $15.732$, compared to $15.432$ for the baseline algorithm.
			
			\item Parent selection, continuous problem set:
			On the Ackley function (\cref{fig-results-individual-parent_selection-cont-0}), $19$ out of $21$ trained agents reached tMBFv values that were better than the baseline algorithm's by a factor of up to $10^6$. The majority of agents improve their MBF near-exponentially across all generations, while the baseline algorithm stagnated after generation $20$. On the Levy \#13 function (\cref{fig-results-individual-parent_selection-cont-2}), $17$ out of $21$ trained agents performed better than the baseline algorithm by up to one order of magnitude. On the Beale (\cref{fig-results-individual-parent_selection-cont-1}) function, the impact of the method was smaller, but many trained agents reached near- or better-than-baseline performance.
			
			\item Survivor selection, continuous problem set:
			Nearly all agents reached better MBFv than the baseline algorithm. On the Ackley (\cref{fig-results-individual-survivor_selection-cont-0}) and Levy \#13 (\cref{fig-results-individual-survivor_selection-cont-2}) function, the baseline algorithm was in many cases outperformed by several orders of magnitude. On the Beale function (\cref{fig-results-individual-survivor_selection-cont-1}), the best agent reached tMBFv that were smaller by a factor of $2$.
		\end{itemize}
		
		\item A minority of the trained agents outperform the baseline algorithm and the variance in performance is large. In the use case with much training time/resources, these methods are still valuable, as one can select the best-performing out of several trained agents. The following adaptation methods pertain to this case:
		
		\begin{itemize}
			
			\item Population-level strategy parameter control:
			Most trained agents performed worse than the baseline algorithm. Nevertheless, on the Ackley (\cref{fig-results-population-learning_parameter-0}) function, a single trained agent achieved a tMBFv that is approximately $10^5$ times better than that of the baseline algorithm. On the Beale (\cref{fig-results-population-learning_parameter-1}) function, two out of $21$ trained agents outperformed the baseline algorithm.
			
			\item Individual-level step-size control:
			This method performed better than the individual-level strategy parameter control method, confirming our idea that eliminating one level of stochasticity by directly controlling step-sizes facilitates the learning of useful policies. On the Ackley (\cref{fig-results-individual-step_size_control-0}) and Beale (\cref{fig-results-individual-step_size_control-1}) function, three out of $21$ trained agents outperformed the baseline algorithm by more than one order of magnitude. On the Levy \#13 function (\cref{fig-results-individual-step_size_control-2}), three agents were able to match its performance.
			
			\item Component-level step-size control (\cref{fig-results-component-step_size_control-cont}):
			On all objective functions, approximately one third of the trained agents outperformed the baseline algorithms, in some cases by multiple orders of magnitude. This is better  than individual-level step-size control, where only one seventh of the trained agents exhibited good performance. These better results are likely due to the method's ability to control mutation along both problem dimensions separately, thus being able to better adapt to the fitness landscape.
			
			\item Component-level binary mutation (\cref{fig-results-component-mutation-knap}):
			Despite the increase in action-space dimensionality, four out of $21$ agents noticeably outperformed the baseline algorithm. The best agent was able to reach a tMBF of $15.931$, compared to the $15.488$ of the baseline algorithm. However, the majority of trained agents were unable to perform any optimization whatsoever  so careful selection of the best agents out of many is particularly important.
		\end{itemize}
		
		\item Only few trained agents match the performance of the baseline algorithm and the variance in performance is large. In this case, static tuning of the parameters of the baseline algorithm is likely more sensible than training many agents just to achieve the same level of performance. The following adaptation methods pertain to this case:
		
		\begin{itemize}
			\item Individual-level mutation rate control (\cref{fig-results-individual-mutation-knap}):
			Only a single trained agent out of $21$ was able to slightly outperform the baseline algorithm (tMBF of $15.538$ compared to $15.488$). A possible explanation is that learning to keep multiple parameters (one per individual) in a very narrow range of feasible values is considerably harder than doing so with a single parameter, as in the population-level method.
			
			\item Individual-level strategy parameter control:
			On the Ackley (\cref{fig-results-individual-learning_parameter-0}) and Beale (\cref{fig-results-individual-learning_parameter-1}) function, only a single trained agent reached a tMBF close to that of the baseline algorithm, exhibiting a faster convergence in the beginning of the optimization process. On the Levy \#13 function (\cref{fig-results-individual-learning_parameter-2}), all trained agents were outperformed by the baseline algorithm. 
		\end{itemize}
	\end{enumerate}

	\section{Conclusions}\label{conclusions}
	The goal of this paper was to investigate whether deep reinforcement learning can be used to improve the effectiveness of evolutionary algorithms and facilitate their application.
	To this end, we developed an approach for learning optimization strategies off-line through deep reinforcement learning.
	
	For experimental evaluation of our approach, we considered use cases in which strategies for previously unseen problem instances have to be learned from a limited set of training instances.
	
	Adaptation methods trained using our approach were in many cases able to outperform classical evolutionary algorithms on combinatorial and continuous optimization tasks.  We also showed that the use of reinforcement learning for evolutionary algorithms is not limited to controlling single numerical parameters of an evolutionary algorithm, but can also be used for both continuous and discrete multi-dimensional control. Furthermore, we achieved promising results with methods that do not merely control existing parameters of evolutionary algorithms, but learn entirely new dynamic fitness functions or selection operators that intelligently guide evolutionary pressure.
	
	However, we noticed that for some of the investigated methods, training was more unstable and results varied more heavily. A more thorough experimental evaluation is required to discern whether this has to be attributed to ill-chosen hyperparameters, the limited size of the used training sets, or the design of the methods.
	Nevertheless, we demonstrated that deep reinforcement learning can be used to improve the effectiveness of evolutionary algorithms.
	
	Further investigation of evolutionary algorithms enhanced by deep reinforcement learning could lead to better population-based optimization algorithms that can more easily be applied to a wide range of problems. 
	To explore the suitability of reinforcement-learning-based adaptation methods to different application domains, future work could consider a wider range of use cases than we did in our experiments, for example:
	\begin{itemize}
		\item unlimited training set (e.g.~problem instances can be randomly generated),
		\item  various degrees of availability of training time/resources,
		\item training to optimize performance on [not necessarily finite] problem instances known at training time (as opposed to generalization to unseen problem instances),
		\item training for multiple problem classes at once (to learn problem-class-independent meta-optimization behavior),
		\item optimization for a variable number of generations.
	\end{itemize}

	Future work should also benchmark against a wider range of methods, and especially 
	combine our approach with a wider range of evolutionary algorithms.

	\section*{Acknowledgments}

	The authors would like to thank Paolo Notaro for valuable discussions.

	\printbibliography

	\appendix[Supplementary Figures and Tables]

	\renewcommand{\thefigure}{S\arabic{figure}}
    \setcounter{figure}{0}

	\renewcommand{\thetable}{S\arabic{table}}
    \setcounter{table}{0}
	
	\label{appendix_figures}

	\begin{figure*}[]
		\centering
		\subfloat[Population-level mutation rate control]{\includegraphics[width=0.32\textwidth]{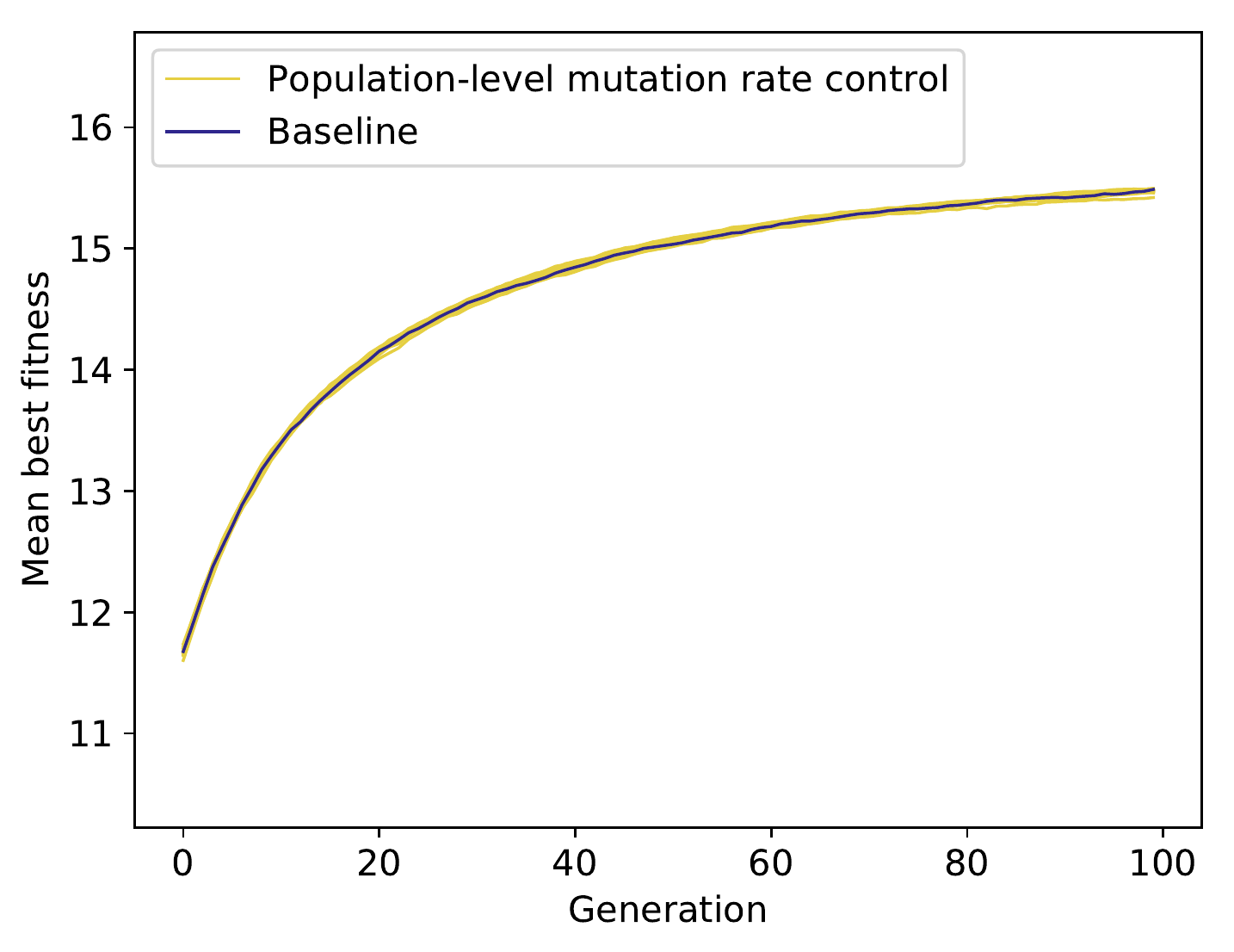}
			\label{fig-results-population-mutation-knap}}
		\hfill
		\subfloat[Individual-level mutation rate control]{\includegraphics[width=0.32\textwidth]{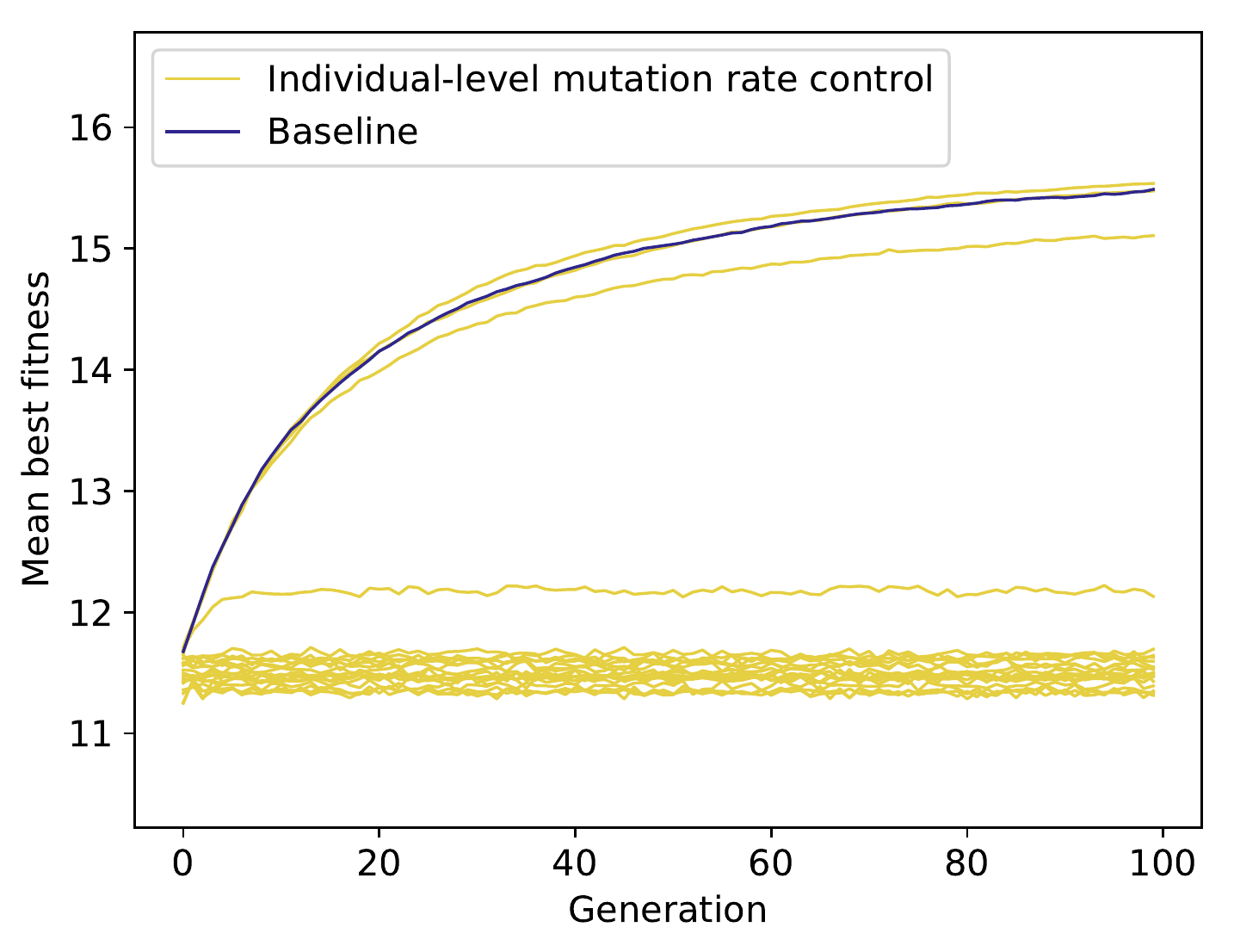}
			\label{fig-results-individual-mutation-knap}}
		\hfill
		\subfloat[Component-level binary mutation]{\includegraphics[width=0.32\textwidth]{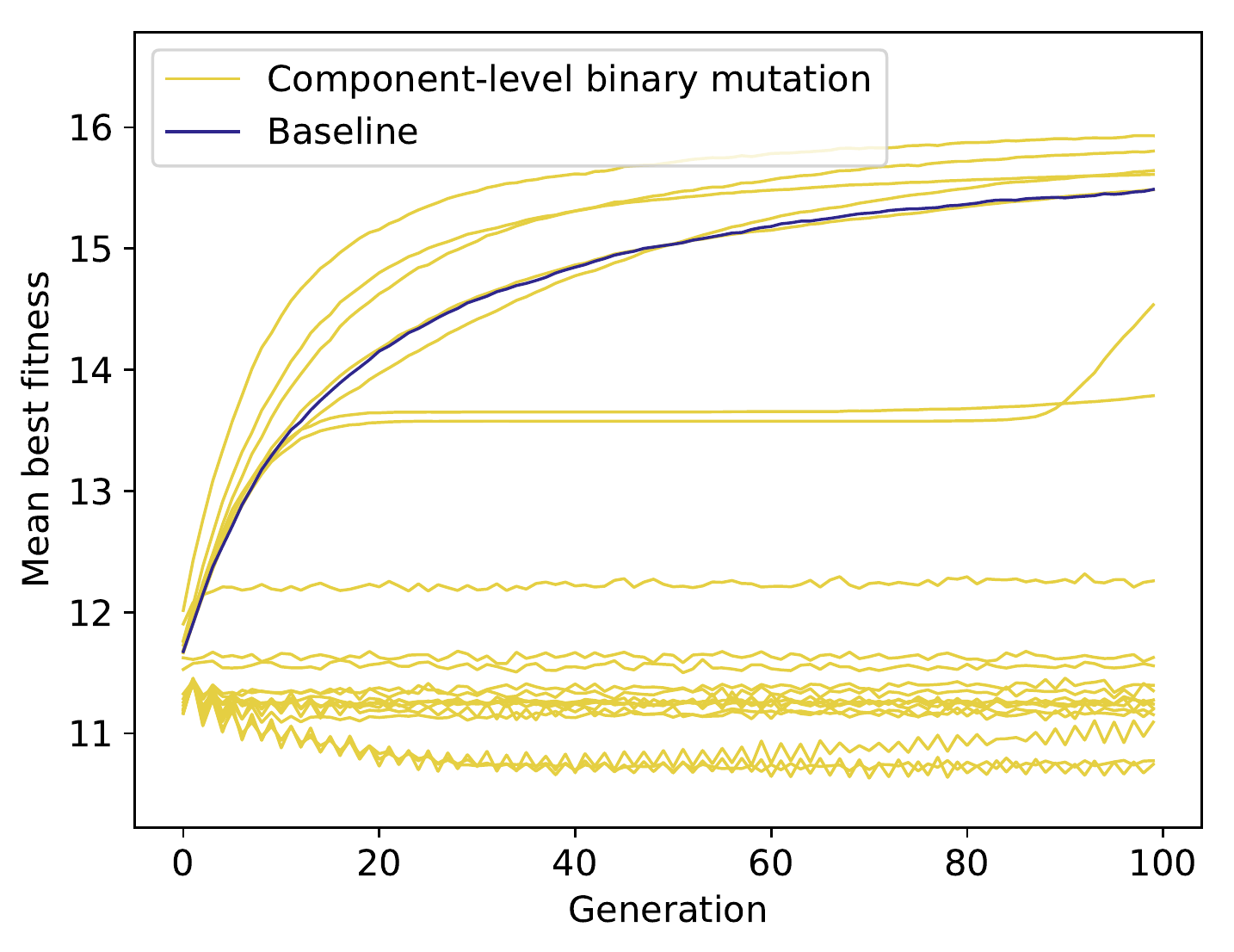}
			\label{fig-results-component-mutation-knap}}

		\caption{MBF (larger is better) of 21 agents trained for the different mutation-controlling adaptation methods (namely on the population level, individual level, and component level) for the knapsack problem, compared to the baseline algorithm. On the population level, all agents matched the performance of the baseline algorithm with an optimized mutation rate. 
		On the individual level, the majority of learned policies lead to a stagnation in fitness. Only one out of $21$ agents  was slightly better than the baseline algorithm.
		On the component level, four trained agents outperformed the baseline algorithm, but the majority of learned policies lead to a stagnation in fitness.
		Given much training time/resources, the user can select the best-performing component-level agent to outperform the baseline algorithm. Component-level methods have outputs with more degrees of freedom, and thus achieve better solutions but are also more difficult to train.}
		\label{fig-results-mutation_rate}
	\end{figure*}
	
	\begin{figure*}[]
		\centering
		\includegraphics[width=0.5\linewidth]{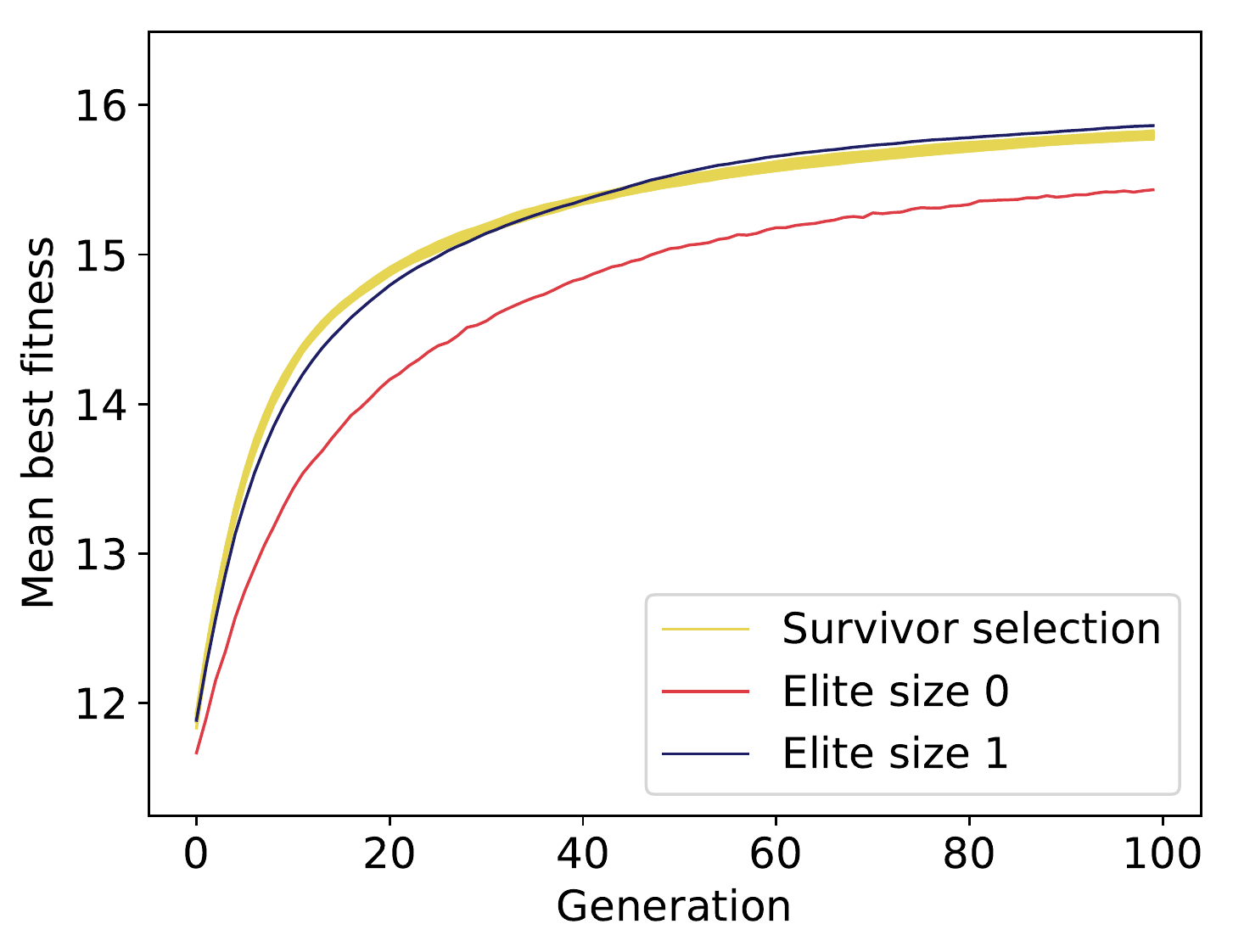}
		\caption{MBF (larger is better) of 21 agents trained for survivor selection on the knapsack problem, compared to the baseline with an optimal elite size of $1$ and an elite size of $0$. The learned policy performed much better than replacement of the population in each generation (i.e.~elite size $0$). Given limited time/resources for training, the user can expect good performance after training a single agent.}
		\label{fig-results-individual-survivor_selection-knap}
	\end{figure*}
	
	\begin{figure*}[]
		\centering
		\subfloat[Fitness shaping]{\includegraphics[width=0.49\textwidth]{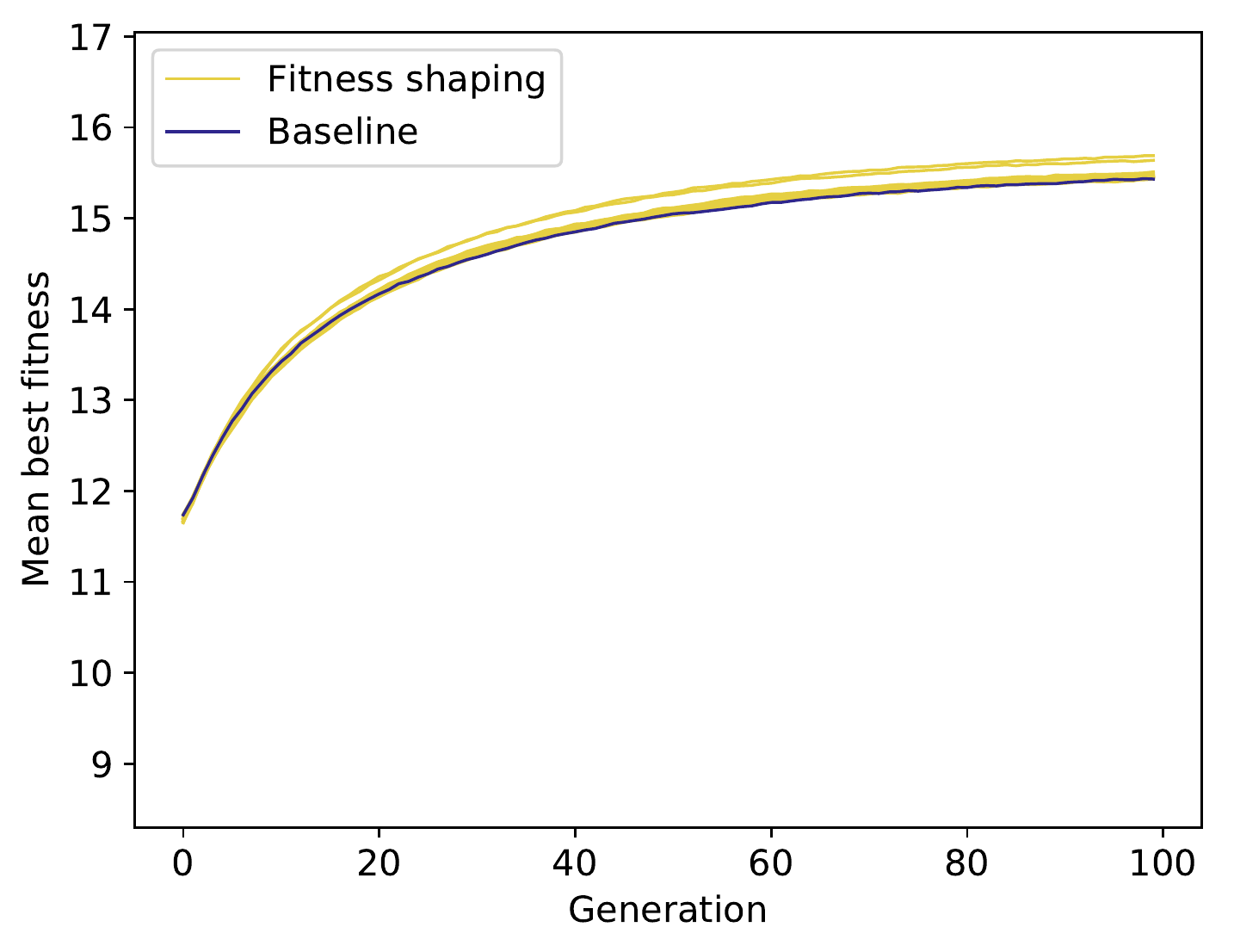}
			\label{fig-results-environment-fitness_shaping-knap}}
		\hfill
		\subfloat[Parent selection]{\includegraphics[width=0.49\textwidth]{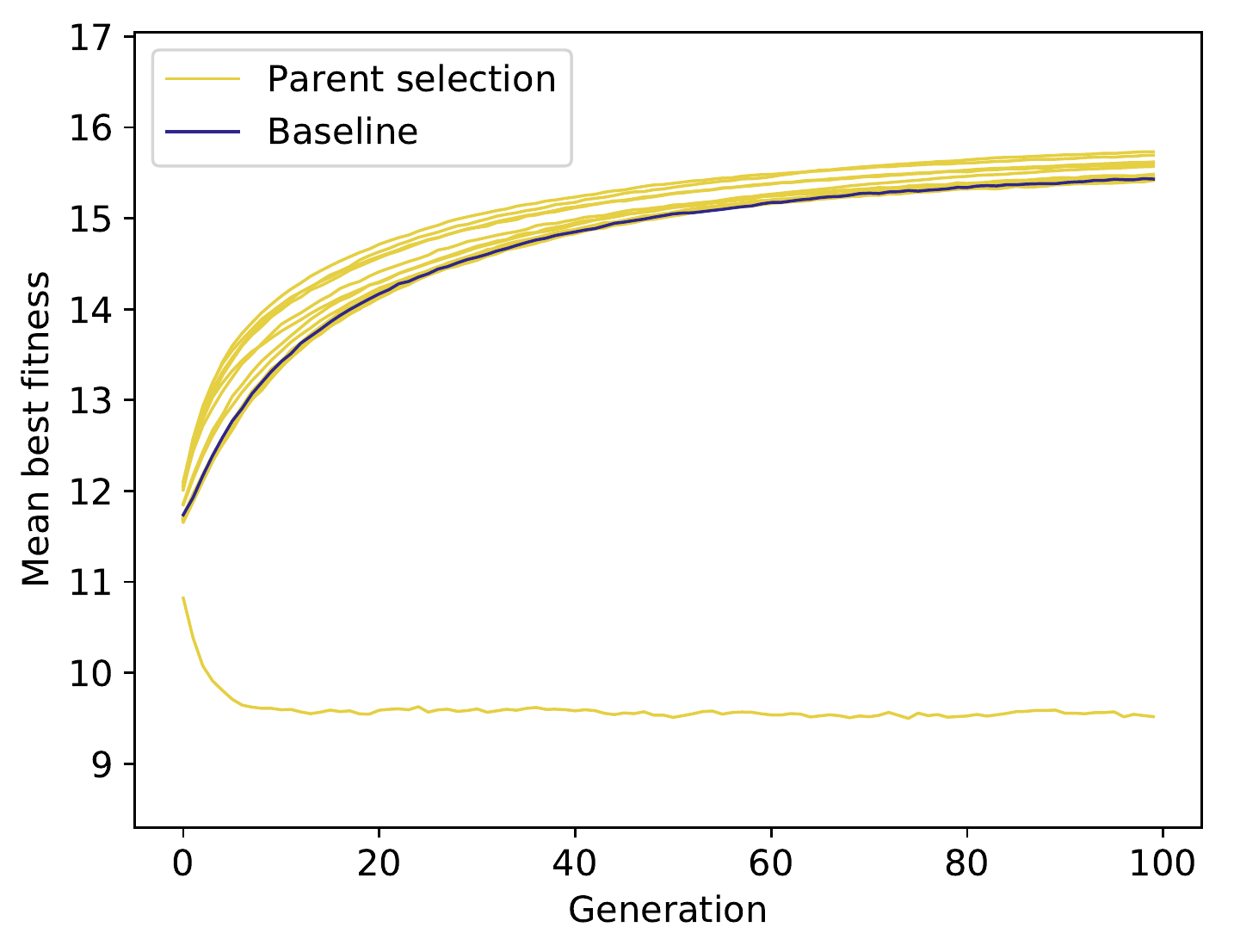}
			\label{fig-results-individual-parent_selection-knap}}

		\caption{MBF (larger is better) of 21 agents trained for parent selection and fitness shaping on the knapsack problem, compared to the baseline algorithm. Except for one outlier, all trained agents of both methods matched the MBF of the baseline algorithm or exceeded it. However, the impact of parent selection is larger, with more agents outperforming the baseline algorithm, most noticeably during the first $20$ generations. With both methods, the user can expect above-baseline performance after a single training run, but choosing the best out of multiple agents is likely to yield even better results.}
		\label{fig-results-parent_selection-fitness_shaping-knap}
	\end{figure*}
	
	\begin{figure*}[]
		\centering
		\subfloat[Optimization of the Ackley function]{\includegraphics[width=0.49\textwidth]{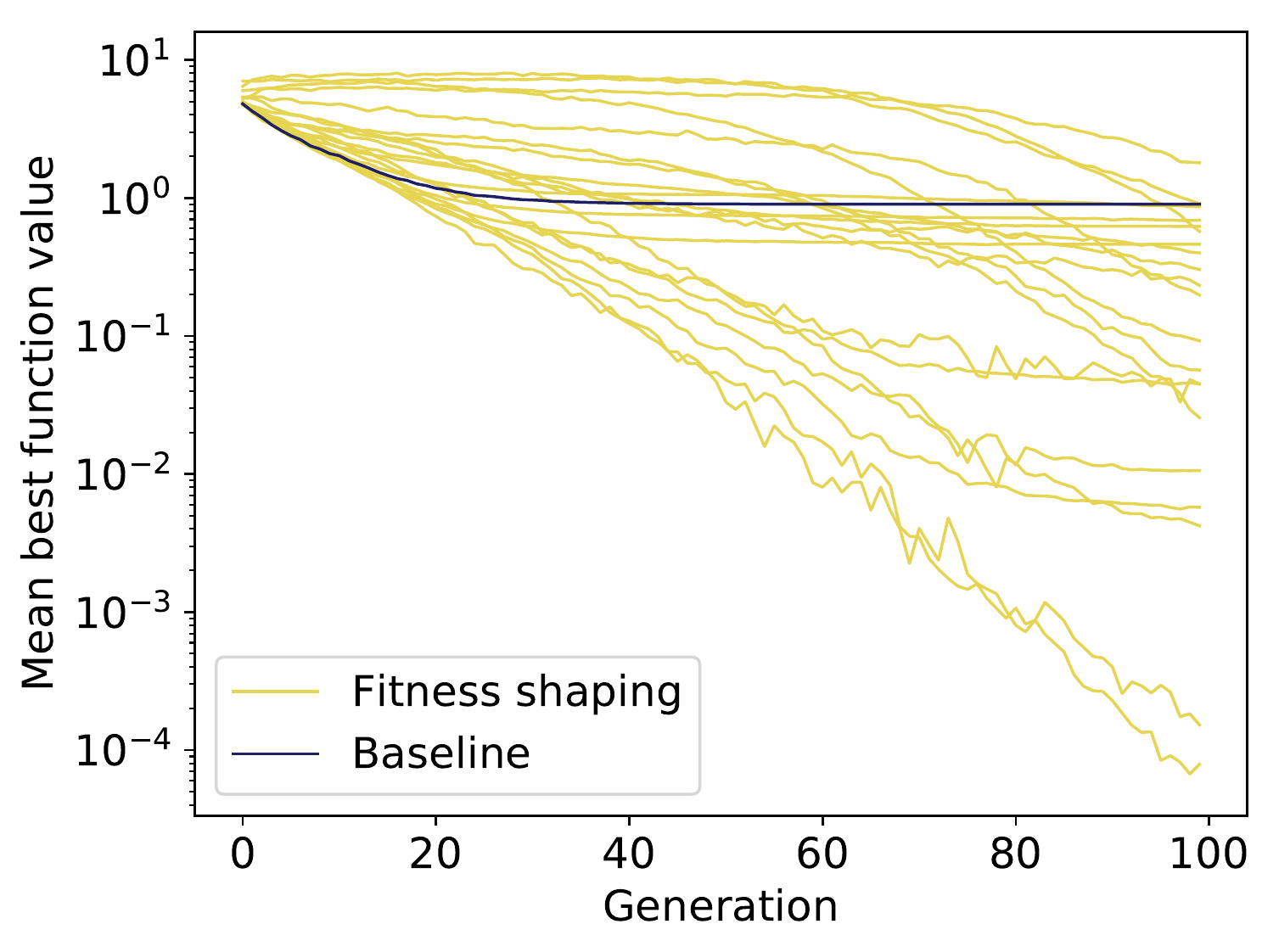}
			\label{fig-results-environment-fitness_shaping-cont-0}}
		\hfill
		\subfloat[Optimization of the Beale function]{\includegraphics[width=0.49\textwidth]{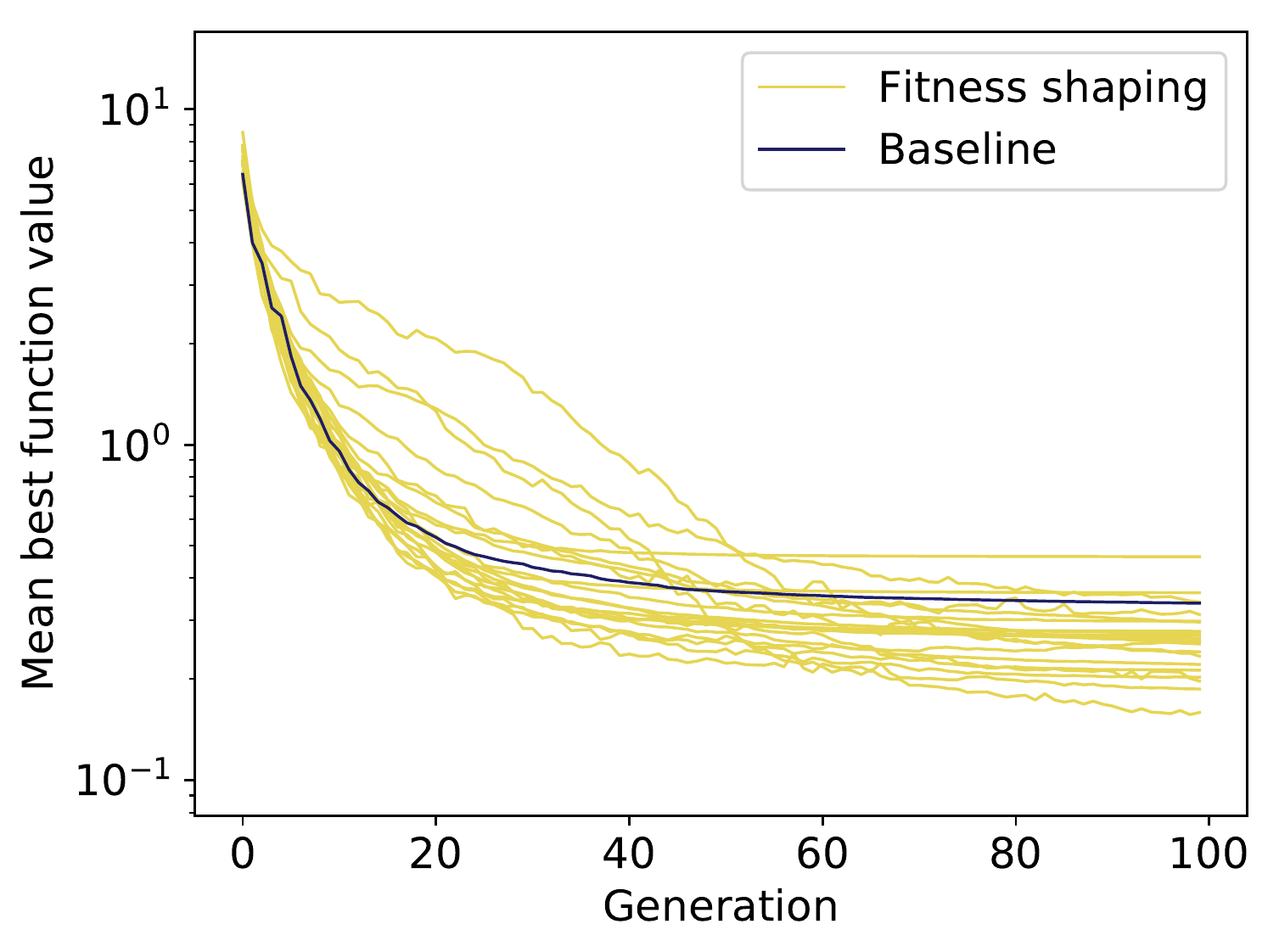}
			\label{fig-results-environment-fitness_shaping-cont-1}}
		
		\caption{. MBFv (smaller is better) of 21 agents trained for fitness shaping, evaluated on the Ackley and Beale functions, compared to the baseline algorithm (see also \cref{fig-results-environment-fitness_shaping-cont-2} for the Levy~\#13 function).
			Nearly all trained agents achieved better-than-baseline performance -- especially on the Ackley function, where the tMBFv of the best trained agent is lower by a factor of more than $10^3$. After a single training run, the user can expect above-baseline performance, but choosing the best out of multiple agents is likely to yield even better results.}
		\label{fig-results-environment-fitness_shaping-cont}
	\end{figure*}
	
	\begin{figure*}[]
		\centering
		\subfloat[Optimization of the Beale function]{\includegraphics[width=0.49\textwidth]{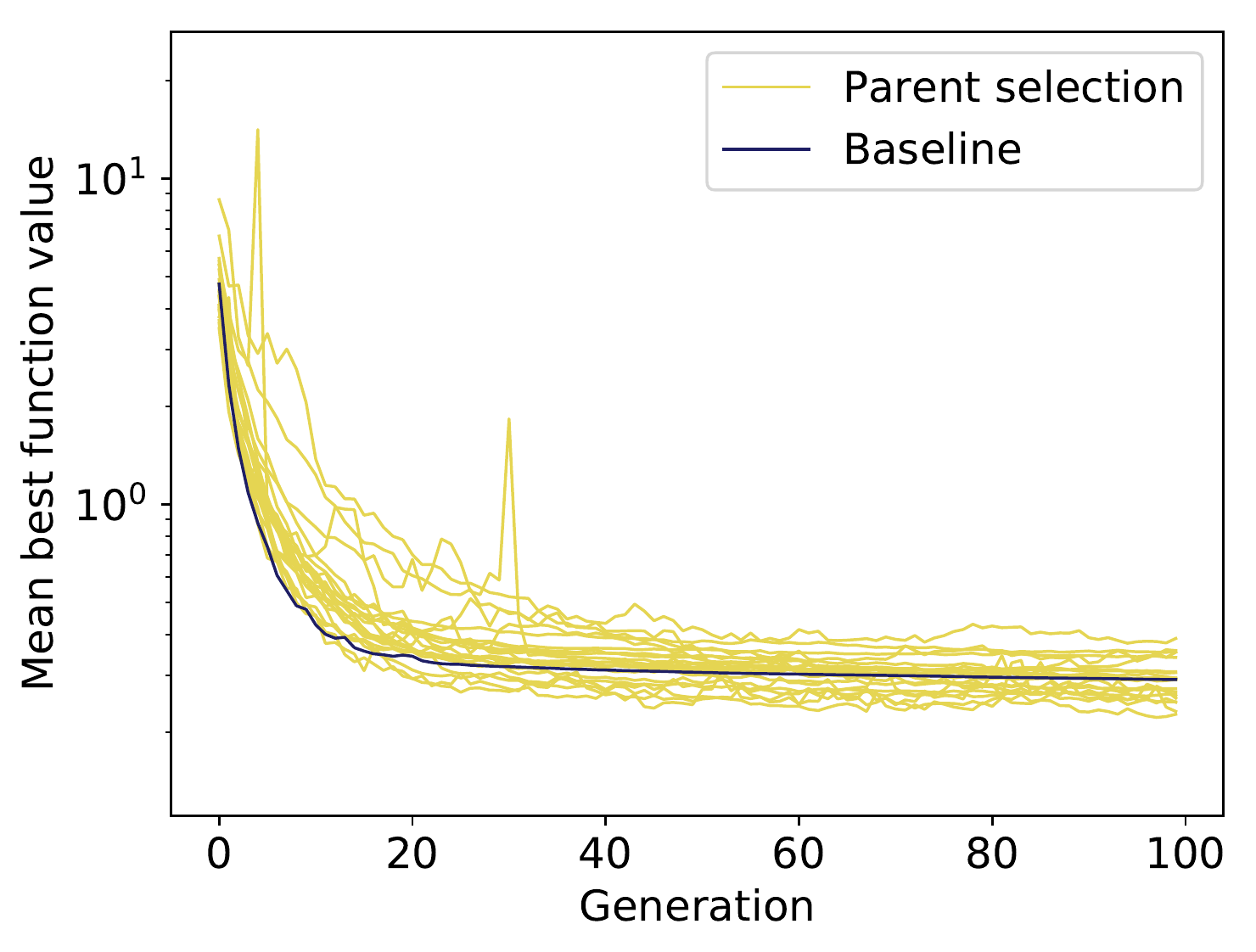}
			\label{fig-results-individual-parent_selection-cont-1}}
		\hfill
		\subfloat[Optimization of the Levy \#13 function]{\includegraphics[width=0.49\textwidth]{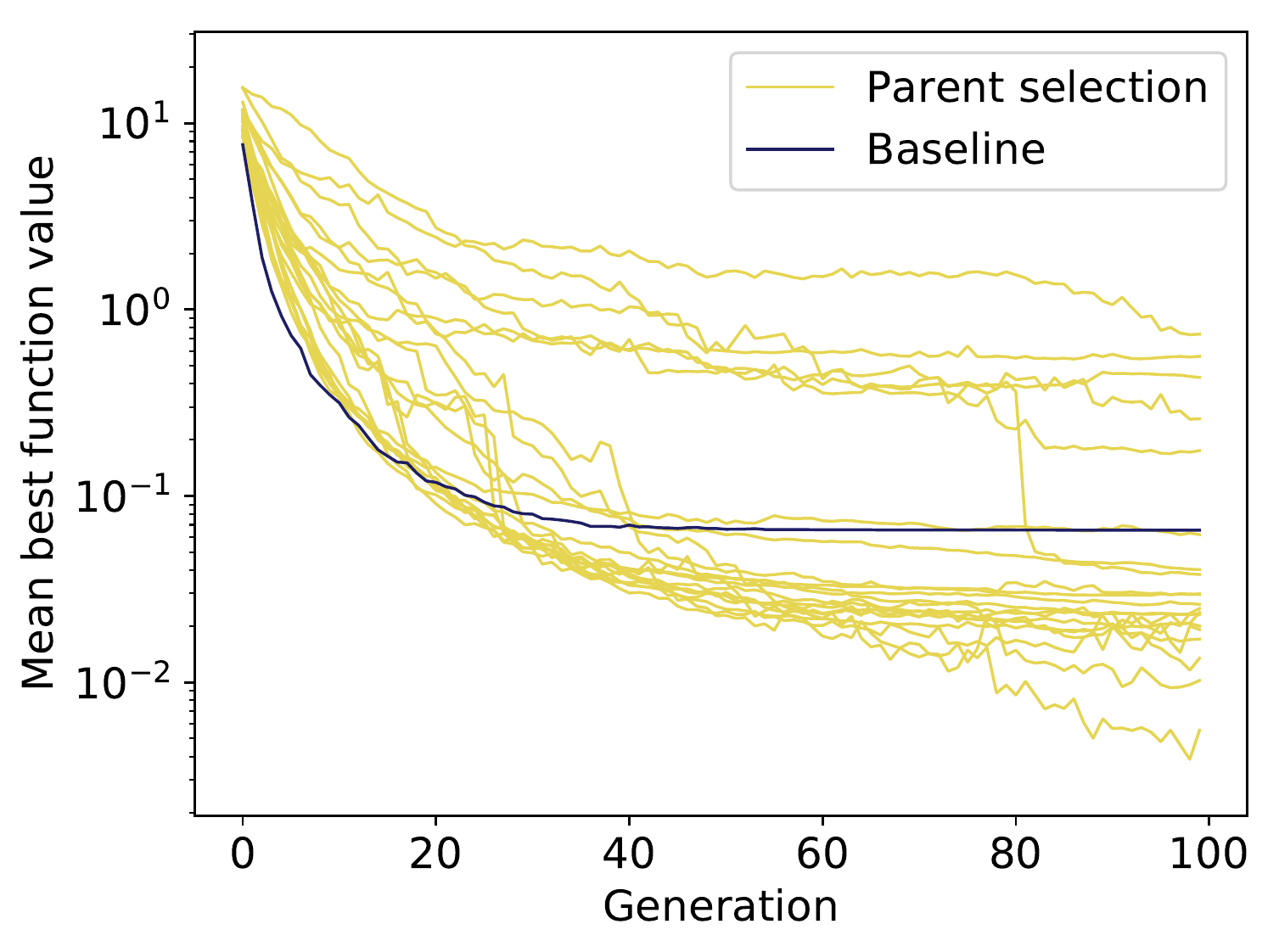}
			\label{fig-results-individual-parent_selection-cont-2}}
		
		\caption{MBFv (smaller is better) of 21 agents trained for parent selection, evaluated on the Beale and Levy \#13 functions, compared to the baseline algorithm (see also \cref{fig-results-individual-parent_selection-cont-0} for the Ackley function). Many of the trained agents outperformed the baseline algorithm on the Levy \#13 function, but tMBFv varies by multiple orders of magnitude among agents. In both methods, the user can expect above-baseline performance after a single training run, but choosing the best out of multiple agents is likely to yield even better results.}
		\label{fig-results-individual-parent_selection-cont}
	\end{figure*}
	
	\begin{figure*}[]
		\subfloat[Optimization of the Beale function]{\includegraphics[width=0.5\textwidth]{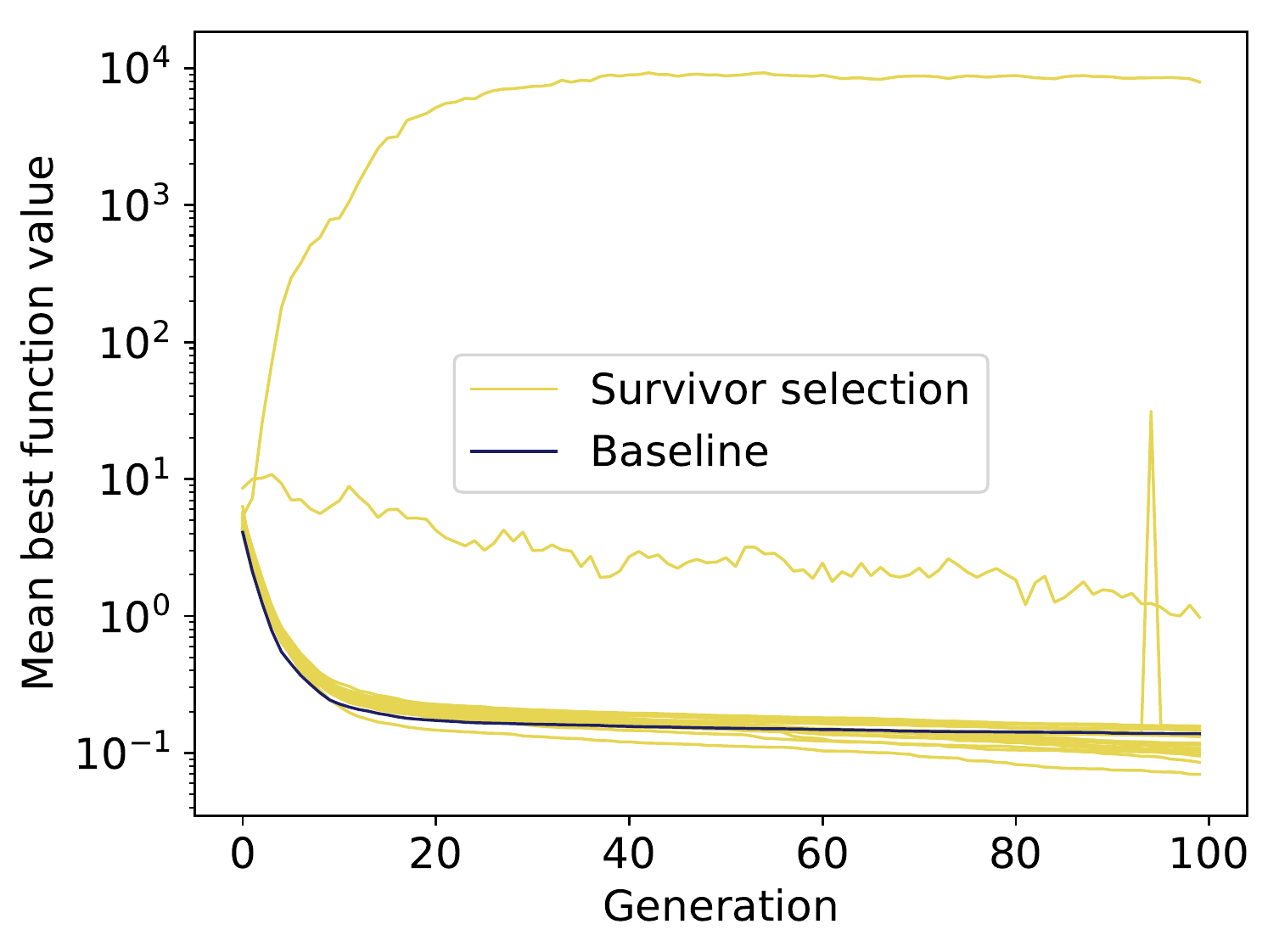}
			\label{fig-results-individual-survivor_selection-cont-1}}
		\hfill
		\subfloat[Optimization of the Levy \#13 function]{\includegraphics[width=0.5\textwidth]{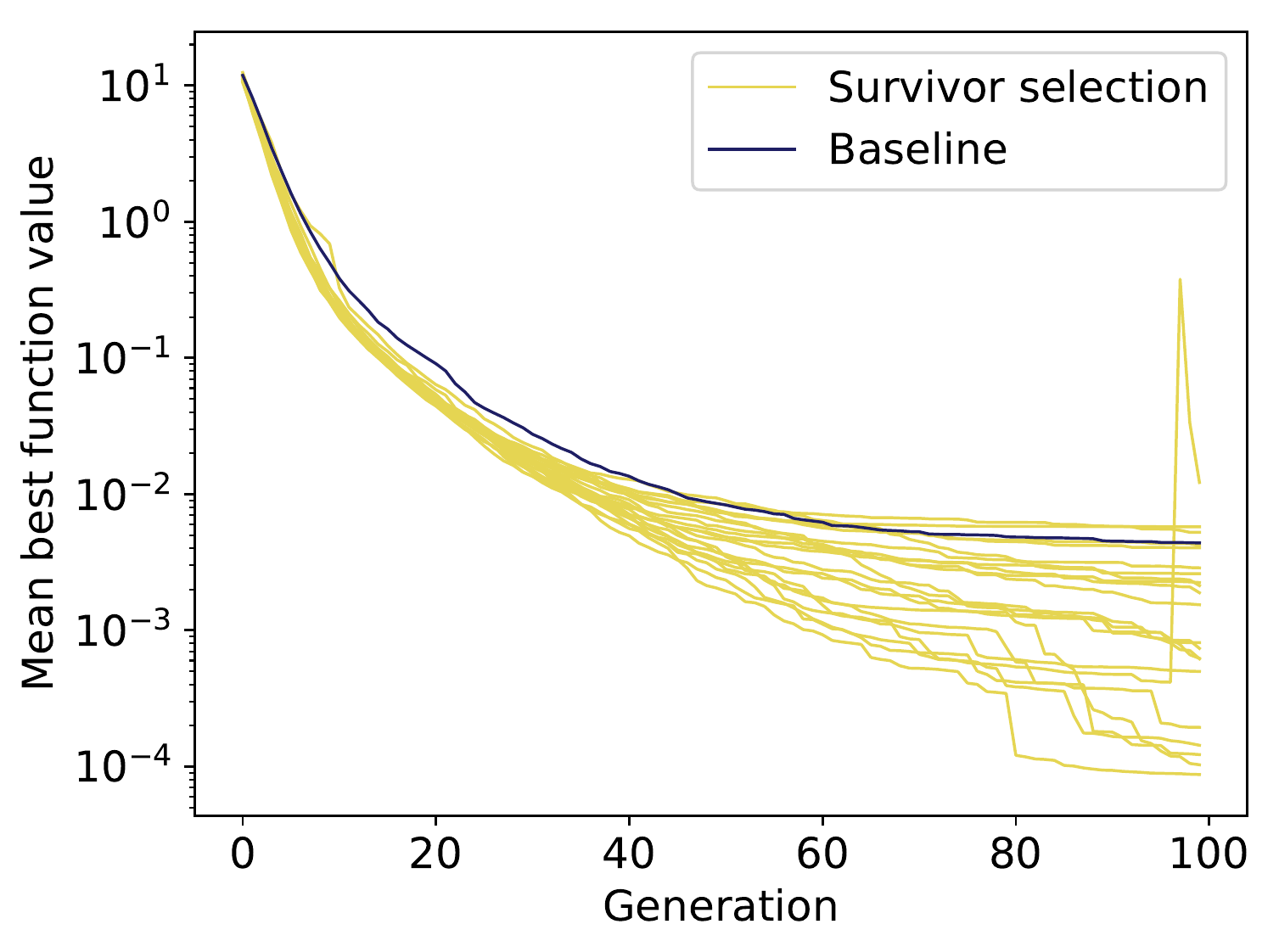}
			\label{fig-results-individual-survivor_selection-cont-2}}
		
		\caption{MBFv (smaller is better) of 21 agents trained for survivor selection, evaluated on the Beale and Levy \#13 functions, compared to the baseline algorithm (see also \cref{fig-results-individual-survivor_selection-cont-0} for the Ackley function). The majority of trained agents performed better than the baseline algorithm, but the variance in tMBFv values among agents was large. After a single training run, the user can expect above-baseline performance, but choosing the best out of multiple agents is likely to yield even better results.}
		\label{fig-results-individual-survivor_selection-cont}
	\end{figure*}
	
	\begin{figure*}[]
		\centering
		\subfloat[Optimization of the Ackley function]{\includegraphics[width=0.32\textwidth]{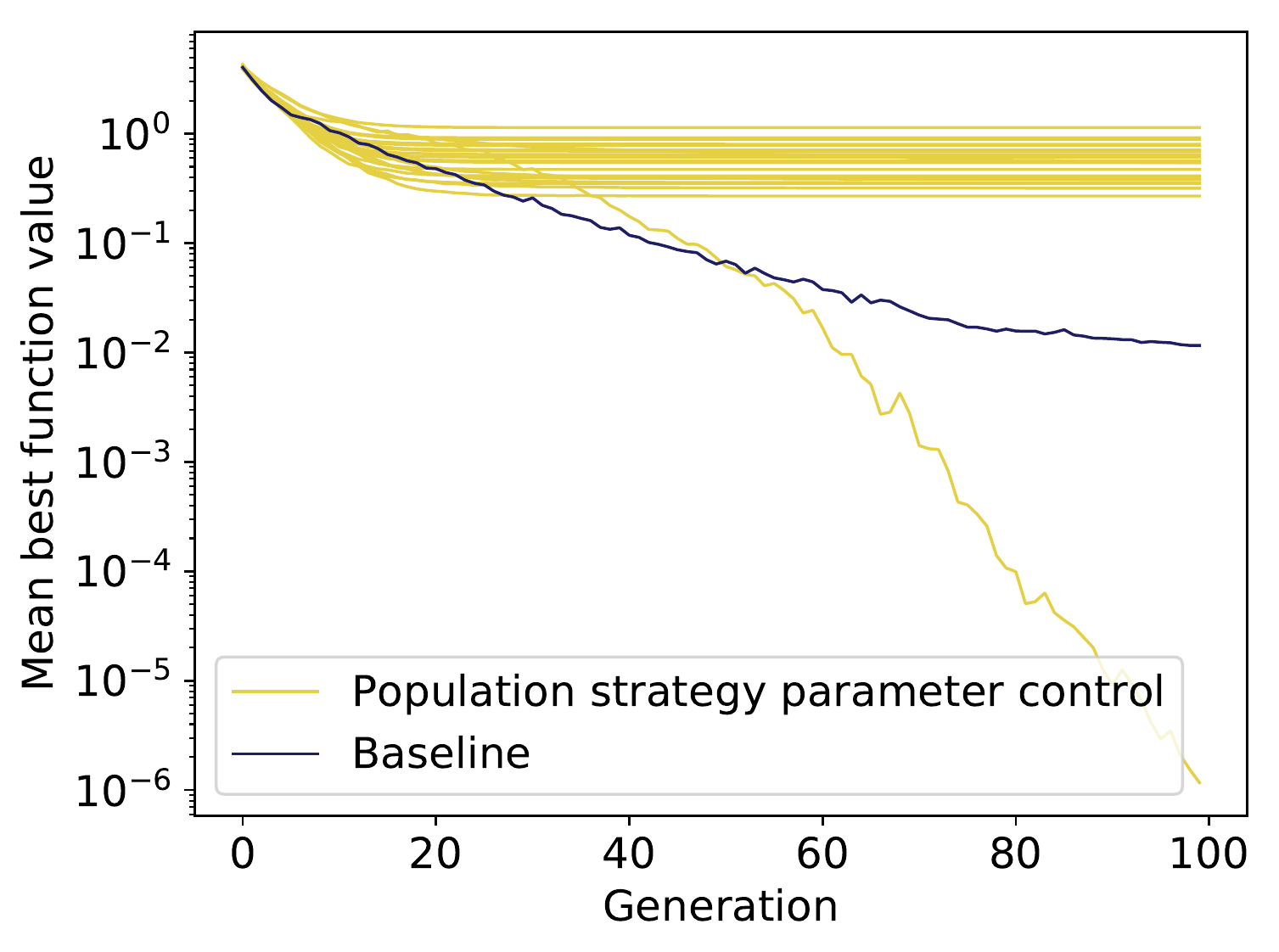}
			\label{fig-results-population-learning_parameter-0}}
		\hfill
		\subfloat[Optimization of the Beale function]{\includegraphics[width=0.32\textwidth]{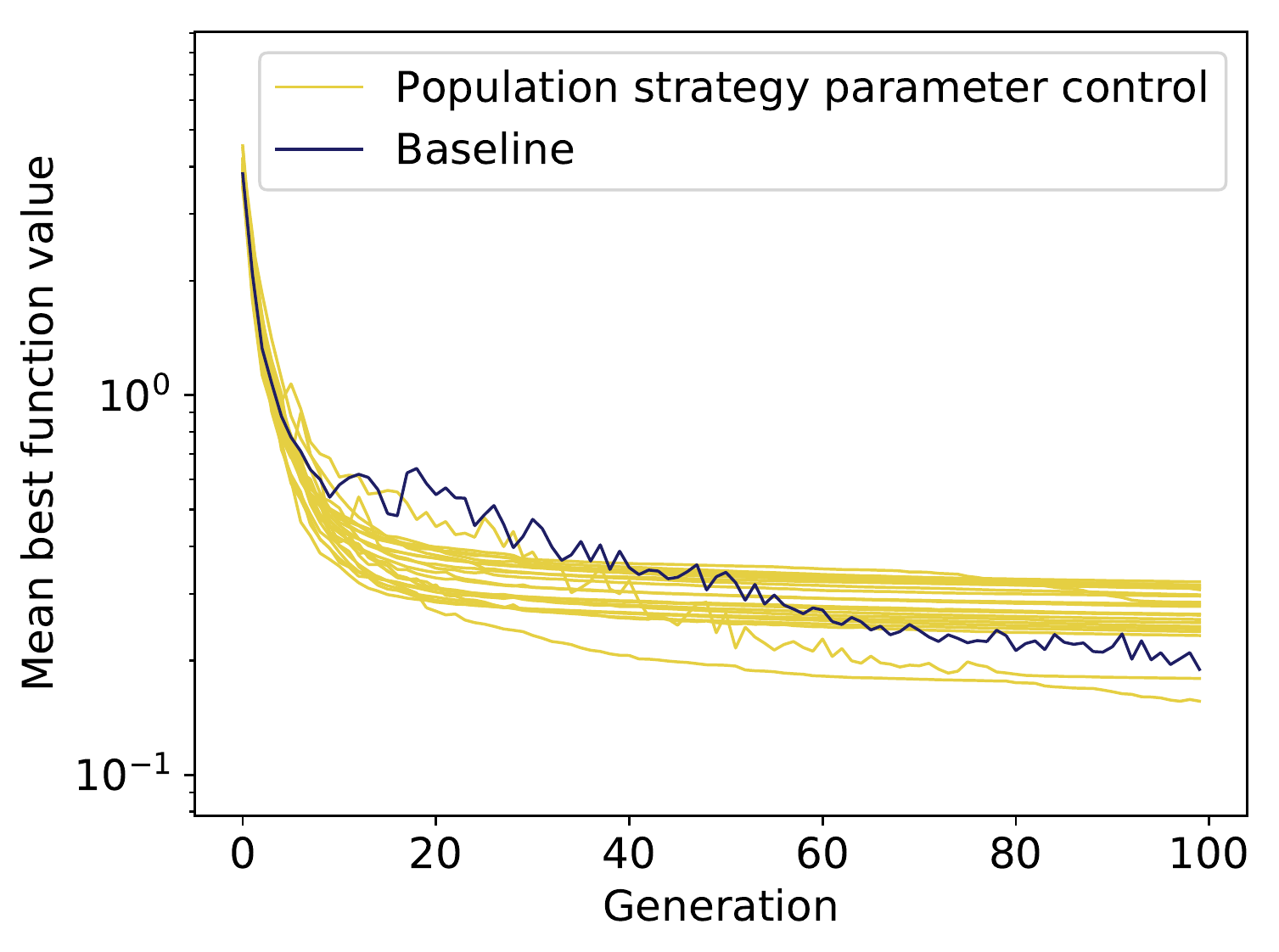}
			\label{fig-results-population-learning_parameter-1}}
		\hfill
		\subfloat[Optimization of the Levy \#13 function]{\includegraphics[width=0.32\textwidth]{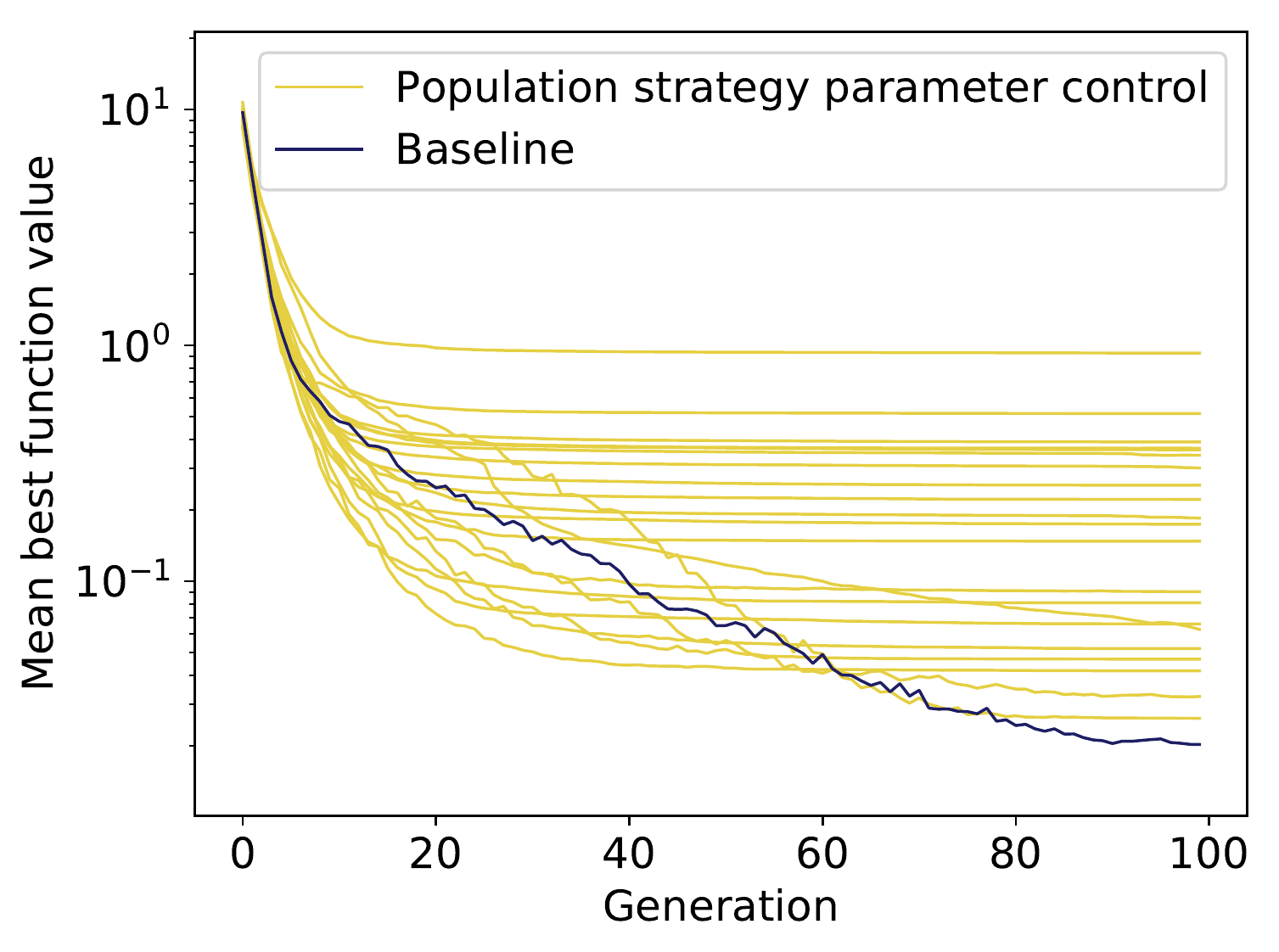}
			\label{fig-results-population-learning_parameter-2}}
		
		\caption{MBFv (smaller is better) of 21 agents trained for population-level strategy parameter control for continuous optimization, evaluated on the validation set, compared to the baseline algorithm. On the Ackley- and Beale function, one and two agents, respectively, outperformed the baseline algorithm. On the Levy \#13 function, two agents reached near-baseline tMBFv values. Although most agents performed worse on all three functions, the best ones performed either better or not much worse than baseline, so that the method could be useful for the use case with much time/resources for training several agents.}
		\label{fig-results-population-learning_parameter}
	\end{figure*}
	
	\begin{figure*}[]
		\centering
		\subfloat[Optimization of the Ackley function]{\includegraphics[width=0.32\textwidth]{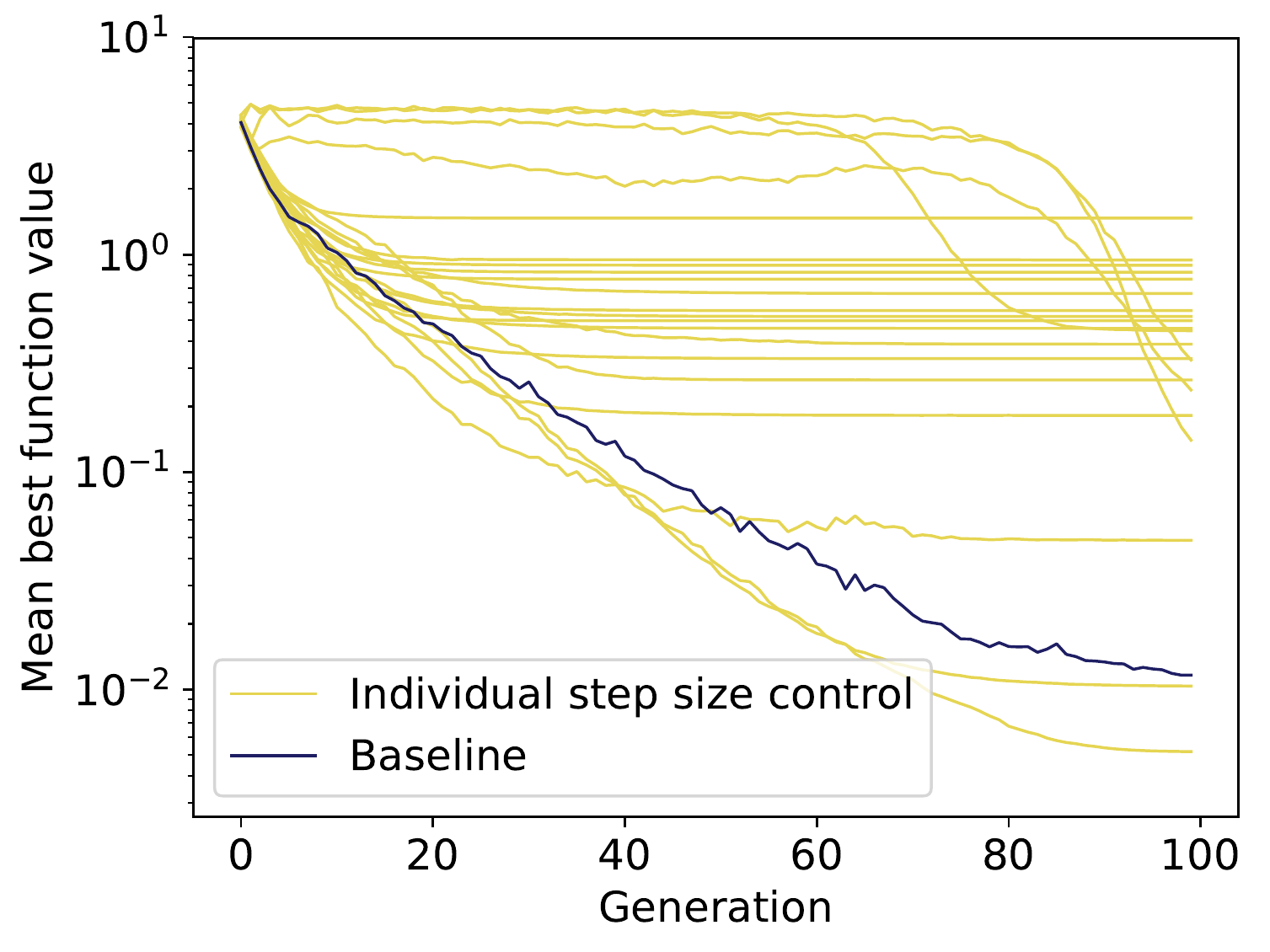}
			\label{fig-results-individual-step_size_control-0}}
		\hfill
		\subfloat[Optimization of the Beale function]{\includegraphics[width=0.32\textwidth]{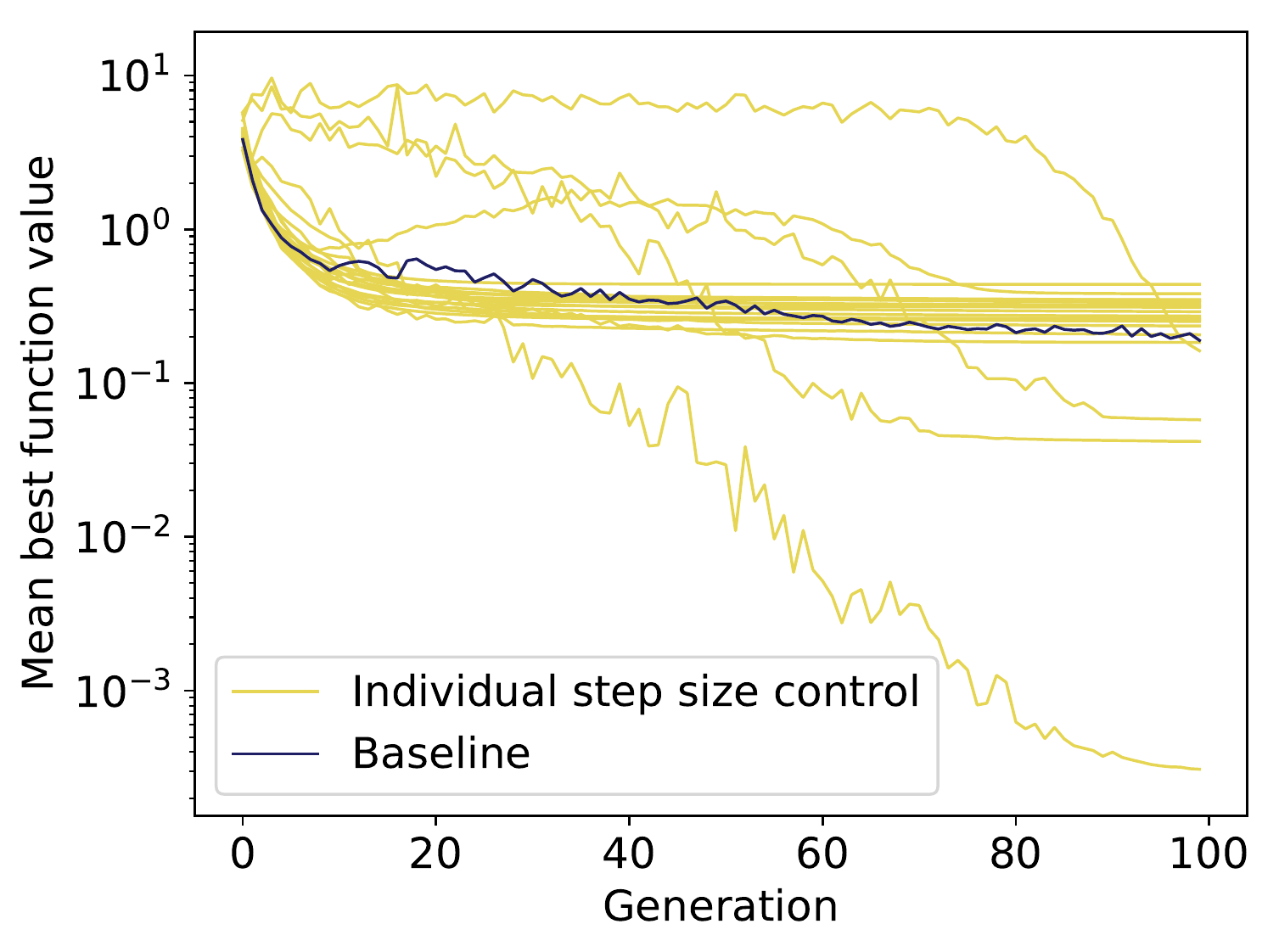}
			\label{fig-results-individual-step_size_control-1}}
		\hfill
		\subfloat[Optimization of the Levy \#13 function]{\includegraphics[width=0.32\textwidth]{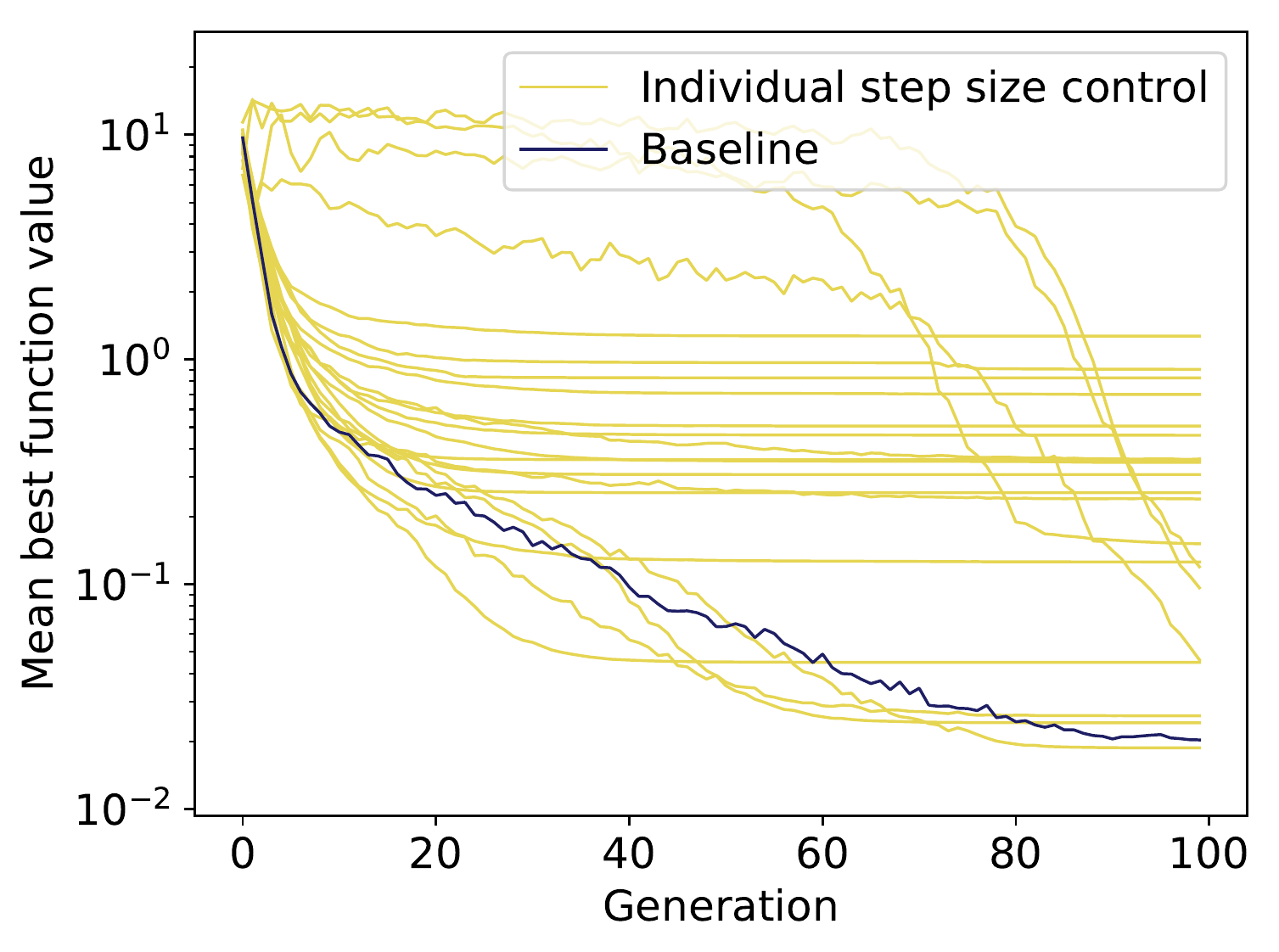}
			\label{fig-results-individual-step_size_control-2}}
		
		\caption{MBFv (smaller is better) of 21 agents trained for individual-level step-size control for continuous optimization, evaluated on the validation set, compared to the baseline algorithm. On the Ackley and Beale function, three out of $21$ trained agents outperformed the baseline algorithm. On the Levy \#13 function, three trained agents achieved baseline levels of performance.
		This adaptation method also performed much better than strategy parameter control (see~\cref{fig-results-individual-learning_parameter}). While both methods control the mutation step-sizes, this method does so deterministically, whereas strategy-parameter control only alters parameters of a random process that changes step-sizes. Reducing this level of stochasticiy likely facilitates training. This method appears well suited for the use case with much training time/resources, in which the user can select the best out of multiple trained agents to improve upon the baseline algorithm.}
		\label{fig-results-individual-step_size_control}
	\end{figure*}
	
	\begin{figure*}[]
		\centering
		\subfloat[Optimization of the Ackley function]{\includegraphics[width=0.32\textwidth]{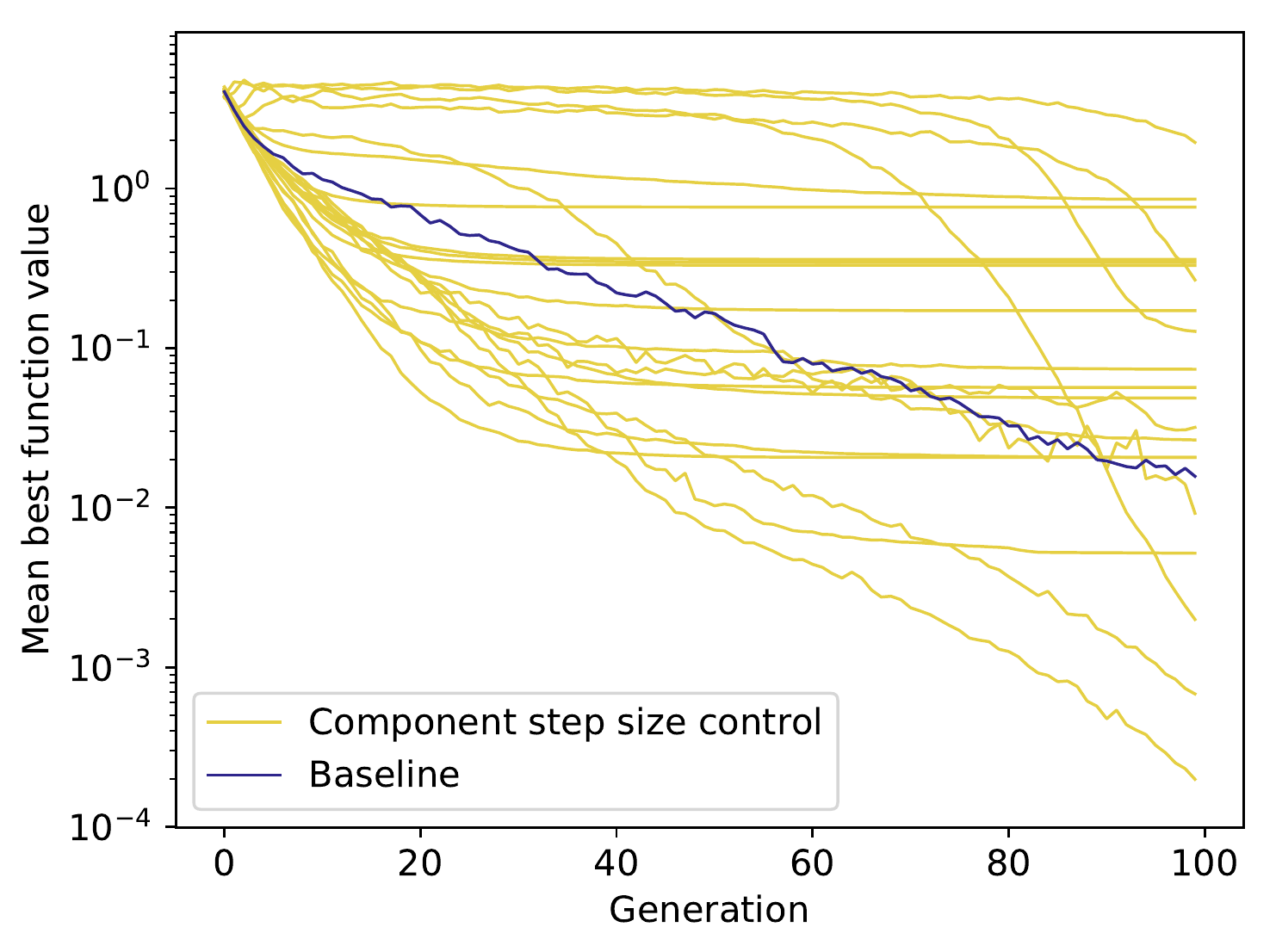}
			\label{fig-results-component-step_size_control-cont-0}}
		\hfill
		\subfloat[Optimization of the Beale function]{\includegraphics[width=0.32\textwidth]{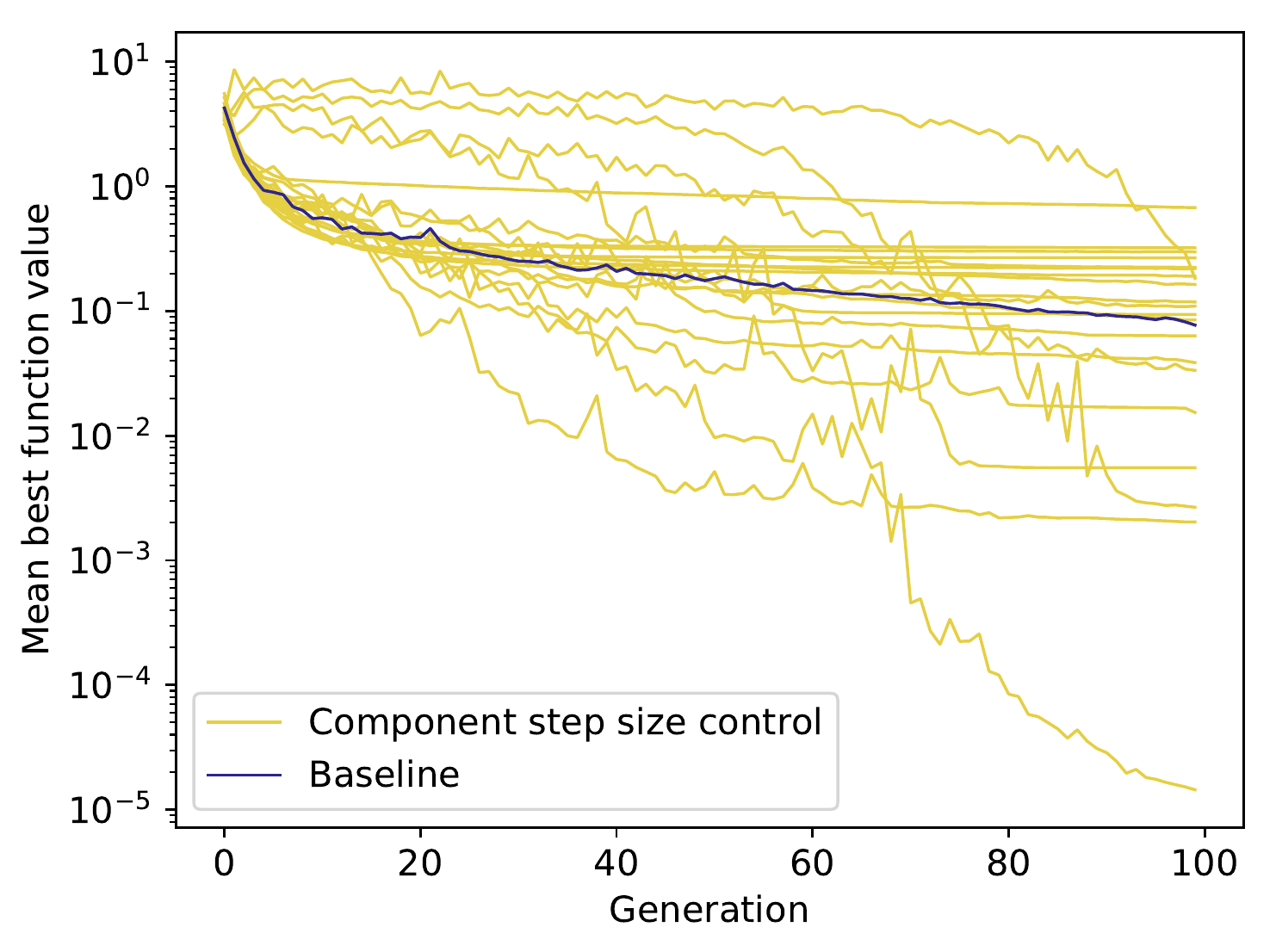}
			\label{fig-results-component-step_size_control-cont-1}}
		\hfill
		\subfloat[Optimization of the Levy \#13 function]{\includegraphics[width=0.32\textwidth]{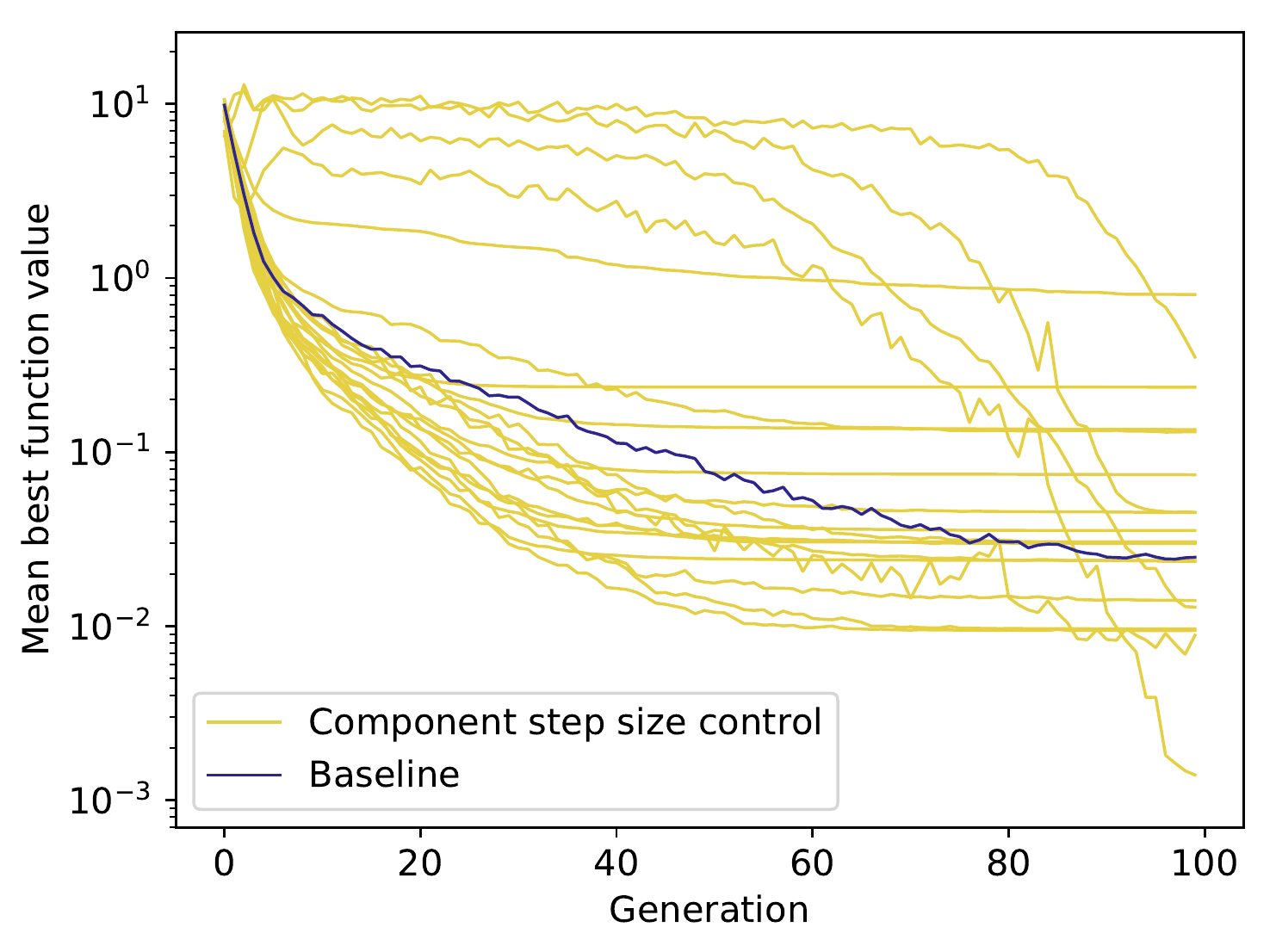}
			\label{fig-results-component-step_size_control-cont-2}}
		
		\caption{MBFv (smaller is better) of 21 agents trained for component-level step-size control for continuous optimization, evaluated on the validation set, compared to the baseline algorithm. Approximately one third of the agents outperformed the baseline algorithm, in many cases by multiple orders of magnitude. This method appears well suited for the use case with much training time/resources, in which the user can select the best out of multiple trained agents to improve upon the baseline algorithm.}
		\label{fig-results-component-step_size_control-cont}
	\end{figure*}
	
	\begin{figure*}[]
		\centering
		\subfloat[Optimization of the Ackley function]{\includegraphics[width=0.32\textwidth]{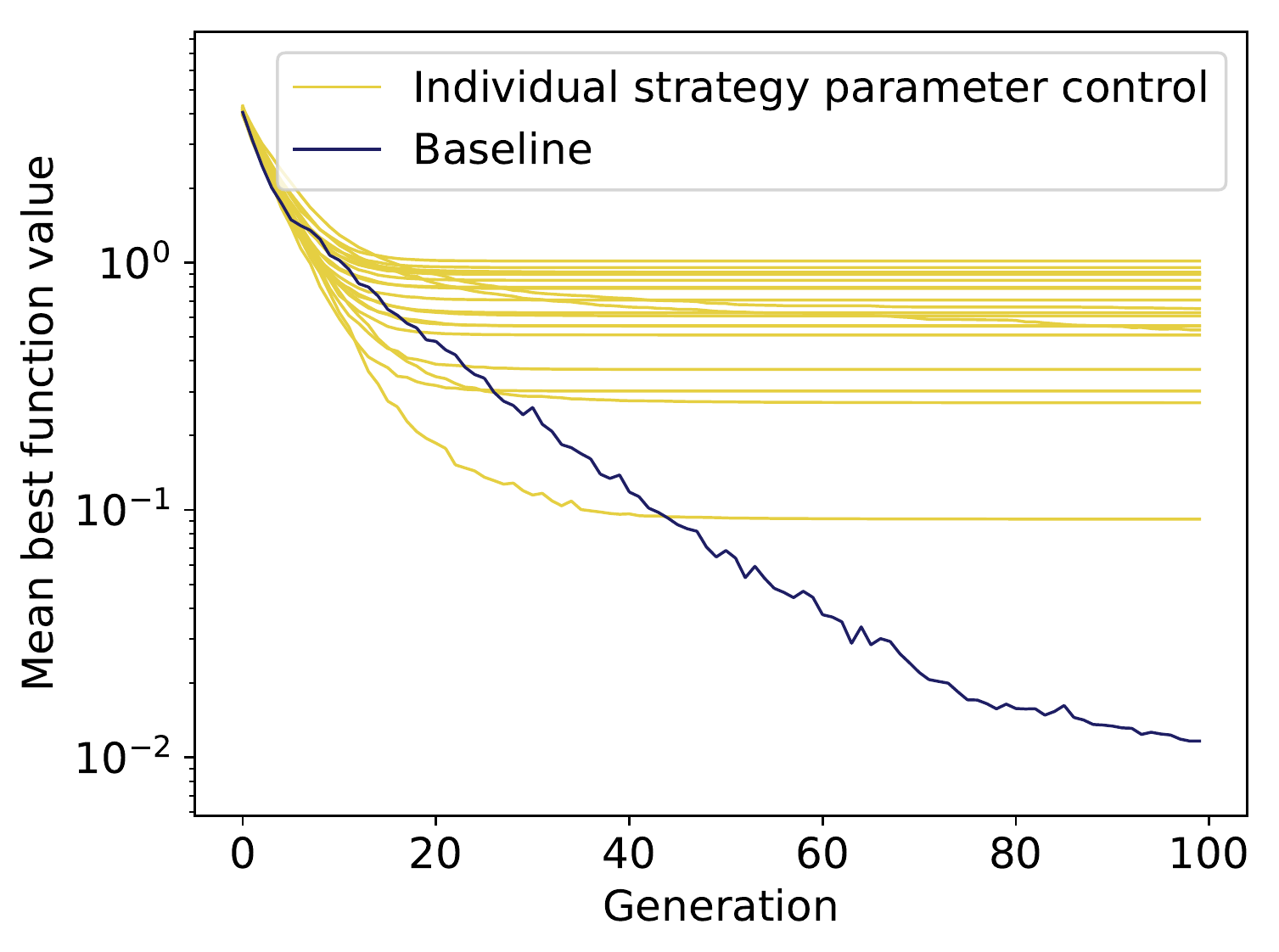}
			\label{fig-results-individual-learning_parameter-0}}
		\hfill
		\subfloat[Optimization of the Beale function]{\includegraphics[width=0.32\textwidth]{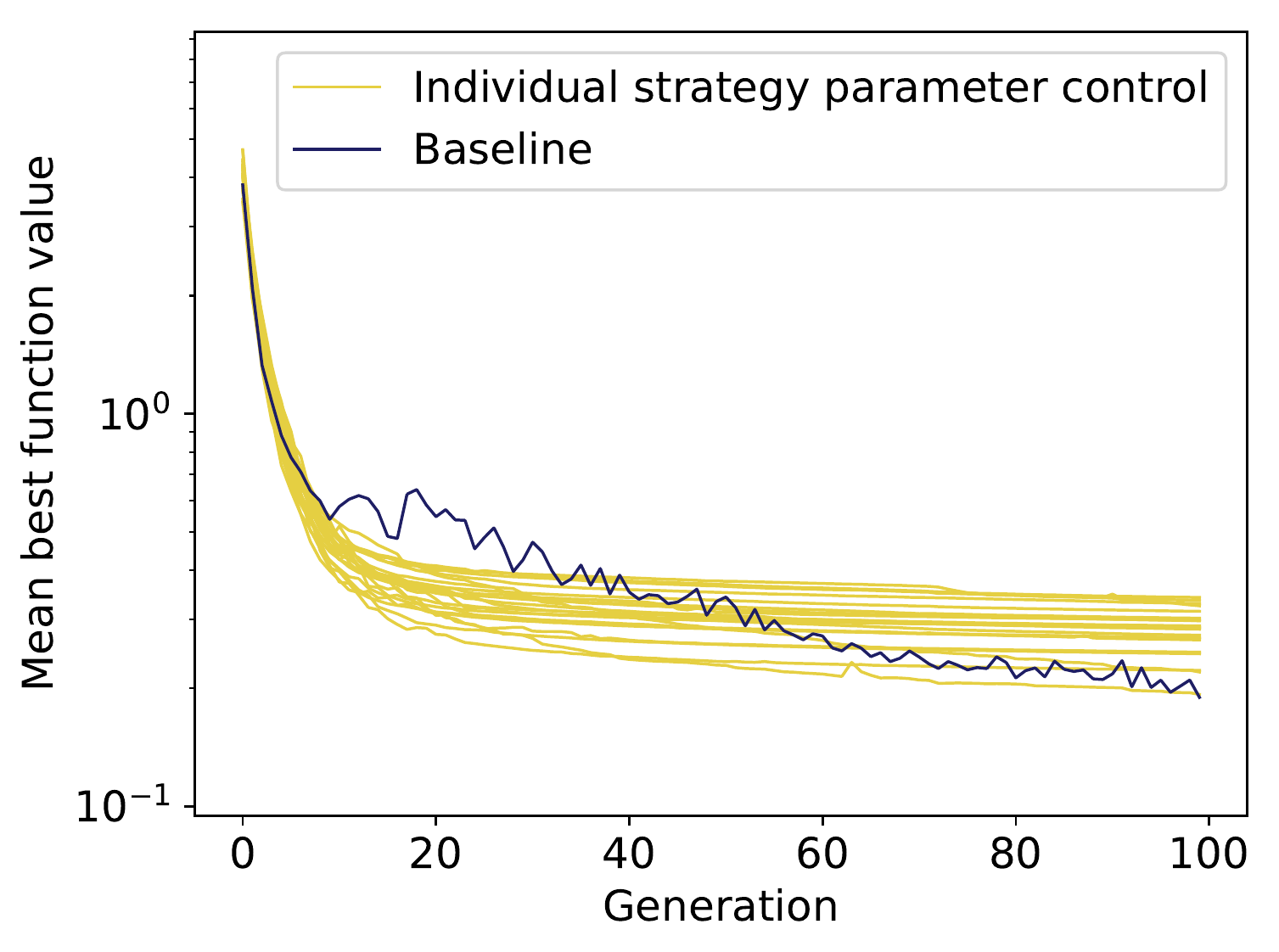}
			\label{fig-results-individual-learning_parameter-1}}
		\hfill
		\subfloat[Optimization of the Levy \#13 function]{\includegraphics[width=0.32\textwidth]{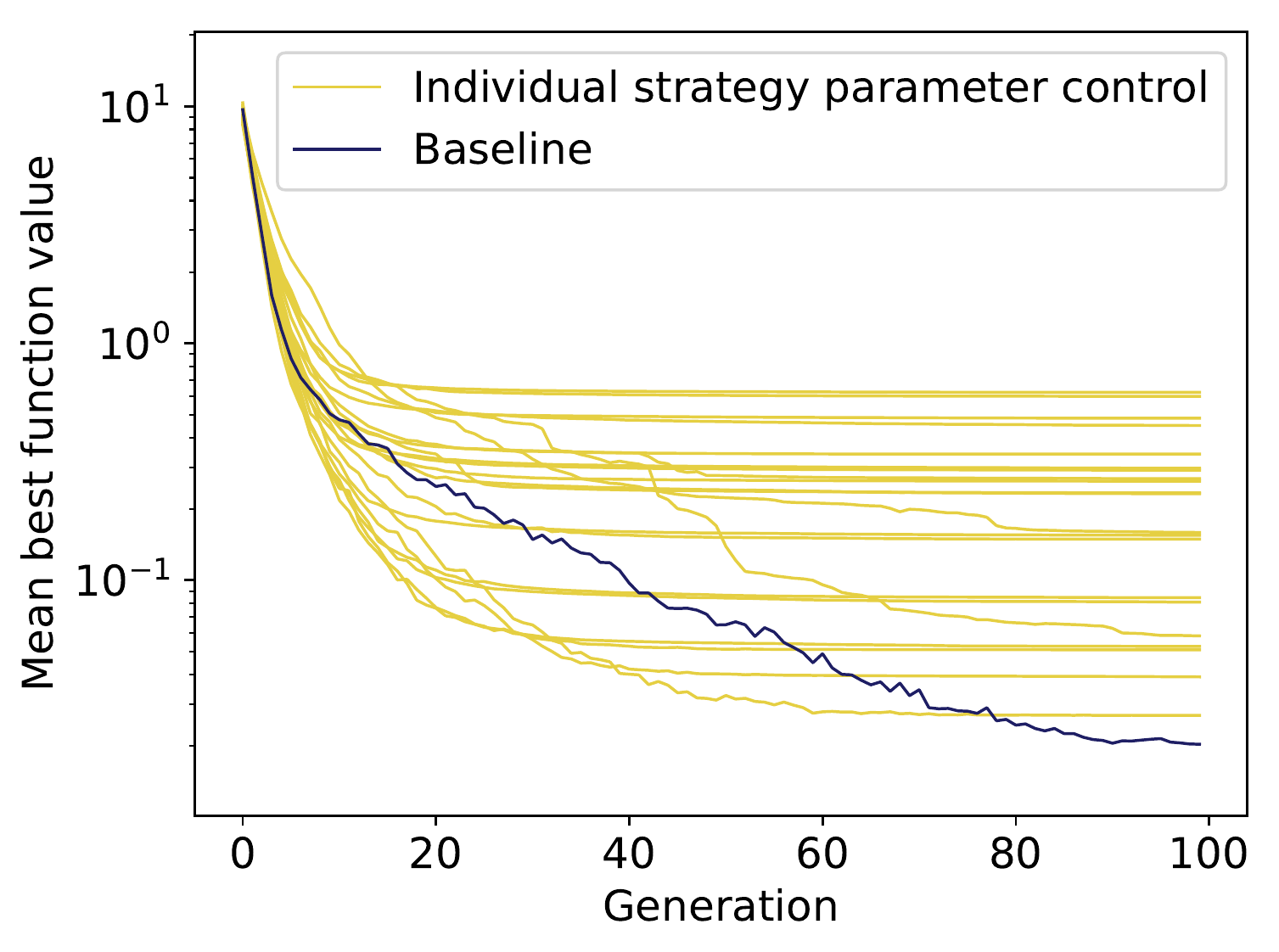}
			\label{fig-results-individual-learning_parameter-2}}
		
		\caption{MBFv (smaller is better) of 21 agents trained for individual-level strategy parameter control for continuous optimization, evaluated on the validation set, compared to the baseline algorithm. Some agents outperformed the baseline algorithm in early generations, but none in late generations.}
		\label{fig-results-individual-learning_parameter}
	\end{figure*}

	\ifCLASSOPTIONcaptionsoff
	\newpage
	\fi
	
	\begin{table*}
	\centering
	\begin{tabular}{| c | c | c |  c | c | c | c | }
 \hline
 Method & Problem & Learning rate & Batch size & \#epochs & $\alpha_e$ & \#actors\\
 \hline
 
 Fitness shaping & Knapsack & $1 \cdot 10^{-4}$ &  $200$ & $8$ & $10^{-4}$ & 4\\
 
 Fitness shaping & Continuous & $5 \cdot 10^{-4}$ & $400$ & $8$ & $10^{-4}$ & 4\\ 
 
 Survivor selection & Knapsack & $1 \cdot 10^{-4}$ &  $400$ & $4$ & $10^{-4}$ & 4\\
 
 Survivor selection & Continuous & $1 \cdot 10^{-4}$ &  $800$ & $8$ & $10^{-4}$ & 4\\
 
 Population-level mutation rate control & Knapsack & $1 \cdot 10^{-4}$ &  $800$ & $4$ & $10^{-4}$ & 4\\
 
 Population-level strategy parameter control & Continuous & $1 \cdot 10^{-4}$ &  $400$ & $4$ & $10^{-4}$ & 4\\
 
 Operator selection & TSP & $1 \cdot 10^{-4}$ &  $400$ & $8$ & $10^{-2}$ & 2\\
 
 Individual-level mutation rate control & Knapsack & $5 \cdot 10^{-4}$ &  $400$ & $4$ & $10^{-4}$ & 4\\
 
 Individual-level strategy parameter control & Continuous & $1 \cdot 10^{-4}$ &  $400$ & $4$ & $10^{-4}$ & 4\\ 
 
 Individual-level step-size control & Continuous & $1 \cdot 10^{-3}$ &  $400$ & $8$ & $10^{-4}$ & 4\\ 
 
 Parent selection & Knapsack & $1 \cdot 10^{-4}$ &  $400$ & $8$ & $10^{-3}$ & 4\\ 
 
 Parent selection & Continuous & $1 \cdot 10^{-4}$ &  $800$ & $8$ & $10^{-3}$ & 4\\ 
 
 Component-level binary mutation & Knapsack & $5 \cdot 10^{-4}$ &  $200$ & $4, 8$ & $10^{-4}$ & 8\\ 
 
 Component-level step-size control & Continuous & $5\cdot 10^{-4}$ &  $800$ & $8$ & $10^{-4}$ & 8\\ 
\hline
\end{tabular}
	
	\caption{Hyperparameter values used for training the different adaptation methods. Some additional hyperparameters were set to the same value across all experiments: $\lambda = \gamma = 0.99$, $\epsilon = 0.2$, $\alpha_v = 0.5$. The reward-scaling factor $\alpha_r$ was set to $100$ for the knapsack and travelling salesman problem and to $1$ for continuous optimization. We used a network depth of $3$ with $64$ filters per layer.}
	\label{deep-learning-hyper_params}
	\end{table*}
	
	\begin{table*}
	\centering
	\begin{tabular}{| c | c | c | c | }
 \hline
 \diagbox{Hyperparameter}{Problem} & Continuous & Knapsack & TSP\\
 \hline
 $\mathrm{population\_size}$ & $10$ & $10$ & $10$\\
 $\mathrm{parent\_percentage}$ & $20\%$, $(50\%)$ \footnotemark[1] & -- & -- \\
 $\mathrm{elite\_size}$ & $0$ & $0$ & $1$\\
 $\mathrm{crossover\_rate}$ & -- & $0.9$ & $1$\\
 $\mathrm{mutation\_rate}$ & -- & 0.01 & 0.01\\
 $\mathrm{strategy\_parameter}$ & $0.5$ & -- & -- \\
 $\mathrm{initial\_step\_size}$ & $0.1$ & -- & -- \\
 $\mathrm{min\_step\_size}$ & $1 \cdot 10^{-8}$ & -- & -- \\
\hline
\end{tabular}
	
	\caption{Default evolutionary parameter values used in benchmarking our adaptation methods on the different problem classes. When evaluating the impact of elite size, the survivor selection adaptation method or the fitness shaping adaptation method, $parent\_percentage$ is set to $50\%$}
	\label{ea-baseline-default-params}
	\end{table*}
	
	\begin{table*}[t!]
	\centering
	\begin{tabular}{| c | c | c | c | c | c | }
 \hline
 \diagbox{Hyper-parameter}{Problem} & Ackley & Beale & Levy \#13 & Knapsack & TSP\\
 \hline
 $\mathrm{parent\_percentage}$ & $20\%$ & $20\%$ & $10\%$ &  -- & -- \\
 $\mathrm{elite\_size}$ & $6$ & $2$ &$9$ & $1$ & -- \\
 $\mathrm{mutation\_rate}$ & -- & -- & -- & 0.0082 & -- \\
 $\mathrm{strategy\_parameter}$ & $0.19$ & $0.11$ & $0.22$ & -- & -- \\
 $\mathrm{crossover\_operator}$ & -- & -- & -- & -- & Two-point \\
\hline
\end{tabular}
	
	\caption{Optimized evolutionary parameter values used in benchmarking our adaptation methods on the different validaton problem sets. Evolutionary algorithms run with these optimized parameter values were compared to corresponding adaptation methods.}
	\label{ea-baseline-optimised-params}
	\end{table*}
	

\end{document}